%% file: main.tex
\documentclass[10pt,twocolumn,letterpaper]{article}

\usepackage[pagenumbers]{cvpr}

\input{preamble}

\definecolor{cvprblue}{rgb}{0.21,0.49,0.74}
\usepackage[pagebackref,breaklinks,colorlinks,allcolors=cvprblue]{hyperref}

\def\confName{CVPR}
\def\confYear{2027}

\title{WorldSmith: Unified Agentic 3D and 4D Content Creation}

\author{%
\textbf{Ke Gui}\textsuperscript{1,2,*}\quad
\textbf{Ru Zhang}\textsuperscript{2,*}\quad
\textbf{Saining Zhang}\textsuperscript{2,3,*,\ensuremath{\dagger}}\quad
\textbf{Weibin Yun}\textsuperscript{2}\\[2pt]
\textbf{Henghaofan Zhang}\textsuperscript{2}\quad
\textbf{Junhao Chen}\textsuperscript{2}\quad
\textbf{Yuanbo Lyu}\textsuperscript{2}\quad
\textbf{Peilin Yang}\textsuperscript{2}\\[2pt]
\textbf{Jiarui Guo}\textsuperscript{2}\quad
\textbf{Jiawei Weng}\textsuperscript{3}\quad
\textbf{Hao Zhao}\textsuperscript{2}\quad
\textbf{Huimin Lu}\textsuperscript{1}\\[4pt]
{\normalsize\textsuperscript{1}National University of Defense Technology}\\[1pt]
{\normalsize
\textsuperscript{2}Tsinghua University\quad
\textsuperscript{3}Nanyang Technological University}\\[5pt]
{\normalsize\urlstyle{same}Project Page: \url{https://gkgo14.github.io/worldsmith/}}
}

\hypersetup{
  pdftitle={WorldSmith: Unified Agentic 3D and 4D Content Creation},
  pdfauthor={Ke Gui, Ru Zhang, Saining Zhang, Weibin Yun, Henghaofan Zhang, Junhao Chen, Yuanbo Lyu, Peilin Yang, Jiarui Guo, Jiawei Weng, Hao Zhao, Huimin Lu}
}

\begin{document}
\twocolumn[{%
\renewcommand\twocolumn[1][]{#1}%
\maketitle
\begin{center}
\vspace{-6pt}
\includegraphics[width=0.92\linewidth]{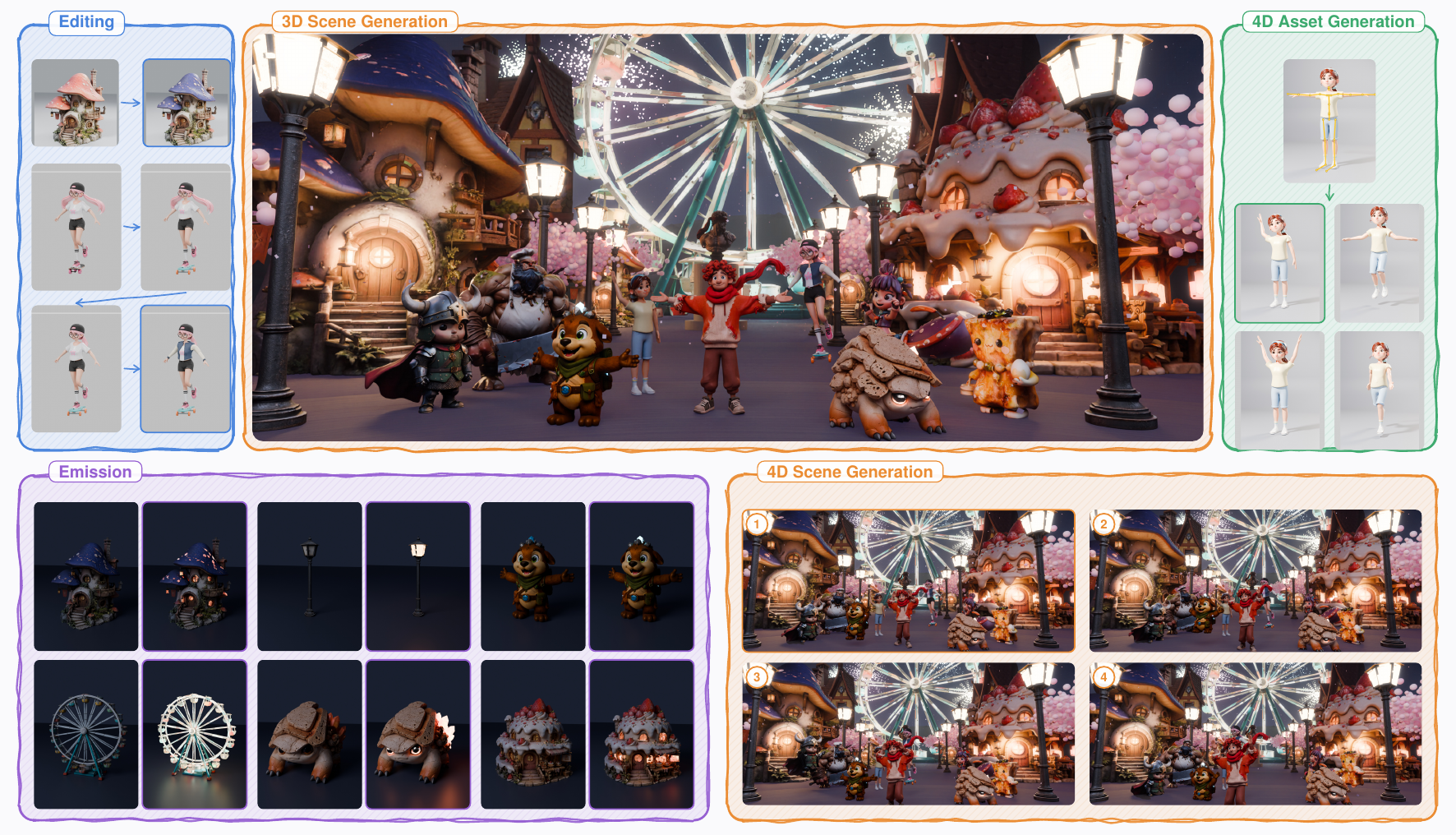}
\vspace{-8pt}
\captionof{figure}{WorldSmith agent for 3D and 4D content. The same agent writes Blender programs, calls foundation models as tools
and checks renders of its own result, without task-specific training. It edits a 3D asset over many turns, writes the
emission of a shape, rigs and animates a character from a video, and reconstructs instance-level 3D and 4D scenes from an
image or a video.}
\label{fig:teaser}
\end{center}%
}]
\begingroup
\renewcommand{\thefootnote}{\fnsymbol{footnote}}
\footnotetext[1]{Equal contribution.}
\footnotetext[2]{Corresponding author.}
\endgroup
\input{sec/0_abstract}
\input{sec/1_intro}
\input{sec/2_related}
\input{sec/3_method}
\input{sec/4_experiments}
\input{sec/5_conclusion}
\raggedbottom   
{
    \small
    \bibliographystyle{ieeenat_fullname}
    \bibliography{main,additional_72}
}

\input{sec/X_suppl}

\end{document}

%% file: preamble.tex
\usepackage{graphicx}
\usepackage{booktabs}
\usepackage{multirow}
\usepackage{xspace}
\usepackage{caption}   

\renewcommand{\dbltopfraction}{0.9}
\renewcommand{\topfraction}{0.9}

\makeatletter
\renewcommand\paragraph{\@startsection{paragraph}{4}{\z@}{1.5ex \@plus .5ex \@minus .2ex}{-1em}{\normalfont\normalsize\bfseries}}
\makeatother

\usepackage{placeins}   

%% file: sec/0_abstract.tex
\begin{abstract}
Creating usable 3D and 4D content requires more than generating geometry: assets must be edited without damaging existing content, articulated motion must respect a rig, and objects must retain their identities when assembled into scenes. These operations are typically addressed by separate models and workflows. We present a unified agentic framework that organizes them around executable Blender programs and visual feedback. A vision-language agent authors code, invokes specialist tools, inspects rendered outcomes, and revises explicit assets within a shared execution environment, without task-specific agent training. Task-specific execution harnesses make this loop concrete: asset editing couples local operations on the original mesh with an issue ledger, candidate-bound evidence, and source-preservation checks; animation operates on native rigs; and scene creation separates layout reasoning from detailed asset generation. The same architecture supports five tasks spanning 3D asset editing, surface emission, rig-based animation, and 3D/4D scene generation. 
Across five tasks, our framework achieves the best reported results on key metrics for editing, emission generation, animation, and scene reconstruction.
\end{abstract}

%% file: sec/1_intro.tex
\section{Introduction}
\label{sec:intro}

Building a usable digital world involves a series of design decisions. An asset may need a local geometric edit, revised material~\cite{add_text2tex,add_texture}, articulated motion~\cite{li2021neuralblendshapes,add_skeletonaware}, and placement that remains coherent with neighboring objects. Each operation must change one part of the asset while preserving the rest. Language-guided 3D editing~\cite{text2mesh,instructnerf2nerf,gaussianeditor,partflow,add_dreameditor,add_sked,add_nerfart}, compositional scene generation~\cite{diffuscene,midi,scenegen,add_atiss,add_sceneformer,add_holodeck,add_layoutgpt}, automatic rigging~\cite{unirig,puppeteer,add_rignet} and dynamic scene synthesis~\cite{fourdgs,cat4d,ovow,add_animate124,add_4dfy} supply strong components for the whole process. Yet composing them still requires deciding what to change, choosing an appropriate tool, and checking its rendered output.

This motivates a question: \emph{can one agent architecture coordinate heterogeneous 3D and 4D creation tasks through a common executable interface?} The challenge is beyond simply producing plausible images. Edits must avoid accumulating damage across turns; motion must be expressed on the actual skeleton; and detailed object generation must respect scene layout. Moreover, a successful tool call does not establish that the intended visual change occurred. The system must connect and coordinate its decisions to observable consequences in the asset it will deliver.

A coding agent that writes graphics programs~\cite{add_3dgpt,add_blenderllm} and inspects the renders takes over part of that coordination. SceneCraft~\cite{scenecraft}
turns scene graphs into Blender programs and improves layouts from
rendered feedback, BlenderGym~\cite{blendergym} benchmarks code-based graphics editing, and VIGA~\cite{viga} interleaves symbolic programs with visual inspection. None of them targets the dominant constraint in asset work: the asset exists, the change is small, and everything outside the changed region has to survive.

We present \emph{WorldSmith}, a unified agent framework that controls an explicit Blender state through code and tool calls. It reads task inputs and rendered candidates, writes changes as Blender Python, and revises  results for five tasks without task-specific training. Segmentation, image editing, mesh generation and rigging enter the same workspace as tool calls, allowing unification to happen at the execution interface. The framework can use different underlying vision-language models without training for any specific task.

The framework couples this shared loop with mechanisms appropriate to each task. Multi-turn editing keeps a prioritized issue ledger tied to the renders that support each entry, while a final check rejects any delivery that altered protected geometry. Animation exposes the native rig, so the agent writes bone and root motion as keyframes and compares candidates against the reference video. Scenes are laid out as low-poly proxies before detailed meshes are fitted, and emission is selected on the texture atlas instead of regenerated. Editing operates on the faces of the original mesh, so the state the agent inspects is the state it delivers.

We evaluate five tasks, from 3D asset editing and surface emission to 4D asset animation and 3D or 4D scene generation. Established benchmarks test local preservation and motion fidelity, and CompoScene-3D and CompoScene-4D add instance-level geometry and motion annotations for composed scenes. On CompoScene-3D, WorldSmith improves object-level AABB IoU from 0.399 to 0.459 over a direct agent using the same model. On CompoScene-4D, it improves object-level AABB IoU from 0.115 to 0.340 over OVOW.

Our contributions are threefold:
\begin{itemize}
    \item A shared executable agent framework for five heterogeneous 3D/4D creation tasks, integrating code-based authoring, specialist tools, and visual inspection without task-specific agent training.
    \item Task-specific mechanisms for controlled revision, including issue-guided local editing with candidate-bound evidence and preservation checks, native-rig animation, and layout-first scene construction.
    \item Evaluation spanning multi-turn editing through compositional dynamic scenes, including two scene benchmarks with instance-level geometry and motion annotations and comparisons against same-model direct agents.
\end{itemize}

%% file: sec/2_related.tex
\section{Related Work}

\subsection{Language-guided 3D editing}

Text2Mesh~\cite{text2mesh} optimizes mesh appearance and local geometry using text
guidance. Instruct-NeRF2NeRF~\cite{instructnerf2nerf} alternates image editing~\cite{add_instructpix2pix} with radiance-field optimization~\cite{add_nerf}, while GaussianEditor~\cite{gaussianeditor} uses semantic tracing and hierarchical Gaussian splatting for controlled scene edits. PartFlow~\cite{partflow} learns feedforward editing from semantic-part transformations with source-aware latent control, and Vinedresser3D~\cite{vinedresser3d} uses a multimodal agent to locate edit regions in a native 3D latent. Program-based editing offers another form of control: ParSEL~\cite{parsel} converts language requests into parameterized shape-editing programs with analytical edit propagation. These methods edit inside the representation they reconstruct, so the delivered mesh is a byproduct of re-optimization or decoding. Our agent edits the faces of the original mesh, which keeps every turn tied to the asset it ships. EditHero~\cite{edithero} supplies the multi-turn setting in which we evaluate this behavior.

Surface emission is appearance prediction on an existing asset. LightGenBench~\cite{lightgenbench} formalizes it as predicting the RGB emission of a
shape from its geometry, its non-emissive material and a thumbnail, and TRELLIS.2-based emission pipelines~\cite{lightgenbench,trellis2} regress the full texture map. We predict a binary selection over texels, reuse base color on the selected ones, and fuse per-view masks on texture atlas, so the emissive region is chosen rather than regenerated.

\subsection{Agentic graphics and scene generation}

The program-and-inspect loop serves both single-asset authoring and whole-scene assembly. SceneCraft~\cite{scenecraft} translates scene graphs into Blender programs and iteratively improves layouts from rendered feedback,
BlenderGym~\cite{blendergym} benchmarks code-based graphics editing and studies the allocation of inference compute to generation and verification, and VIGA~\cite{viga}, a particularly close work, interleaves symbolic programs with
visual inspection across reconstruction, multi-step editing and physical interaction. Scene generation instead produces all instances at once. DiffuScene~\cite{diffuscene} models scenes as collections of object attributes with
denoising diffusion, MIDI~\cite{midi} jointly models multiple instances conditioned on an image, and SceneGen~\cite{scenegen} predicts a scene in one feedforward pass.

We keep the agent and change the order, fixing a low-poly layout for each object before its detailed geometry is generated. The resulting scene retains separate object instances that can be animated in the 4D pipeline.

\subsection{Rigging, animation and dynamic scenes}

UniRig~\cite{unirig} generates skeletons and skinning weights through autoregressive modeling and bone-point attention. Puppeteer~\cite{puppeteer} addresses rigging and animation together. Our animation stage takes the native rig as its control interface rather than predicting motion in a learned space, and checks candidates against the reference video in paired previews. For dynamic content~\cite{add_dnerf,add_nerfies,add_hypernerf,add_dynamicgaussians,add_hexplane,add_kplanes,huang2026arthoi,shen2026gaussianart,wang2025bilateraldriving}, 4D Gaussian Splatting~\cite{fourdgs} models time-varying appearance and geometry for efficient rendering, CAT4D~\cite{cat4d} combines multi-view video diffusion with deformable Gaussian reconstruction, and OVOW~\cite{ovow} targets physical 4D scenes from monocular video. Our 4D output keeps separate scene objects and a rigged character with explicit animation, so scene construction and motion authoring compose inside one workflow and stay editable.

%% file: sec/3_method.tex
\section{Method}
\label{sec:method}
\begin{figure*}[t]
  \centering
  \includegraphics[width=\textwidth]{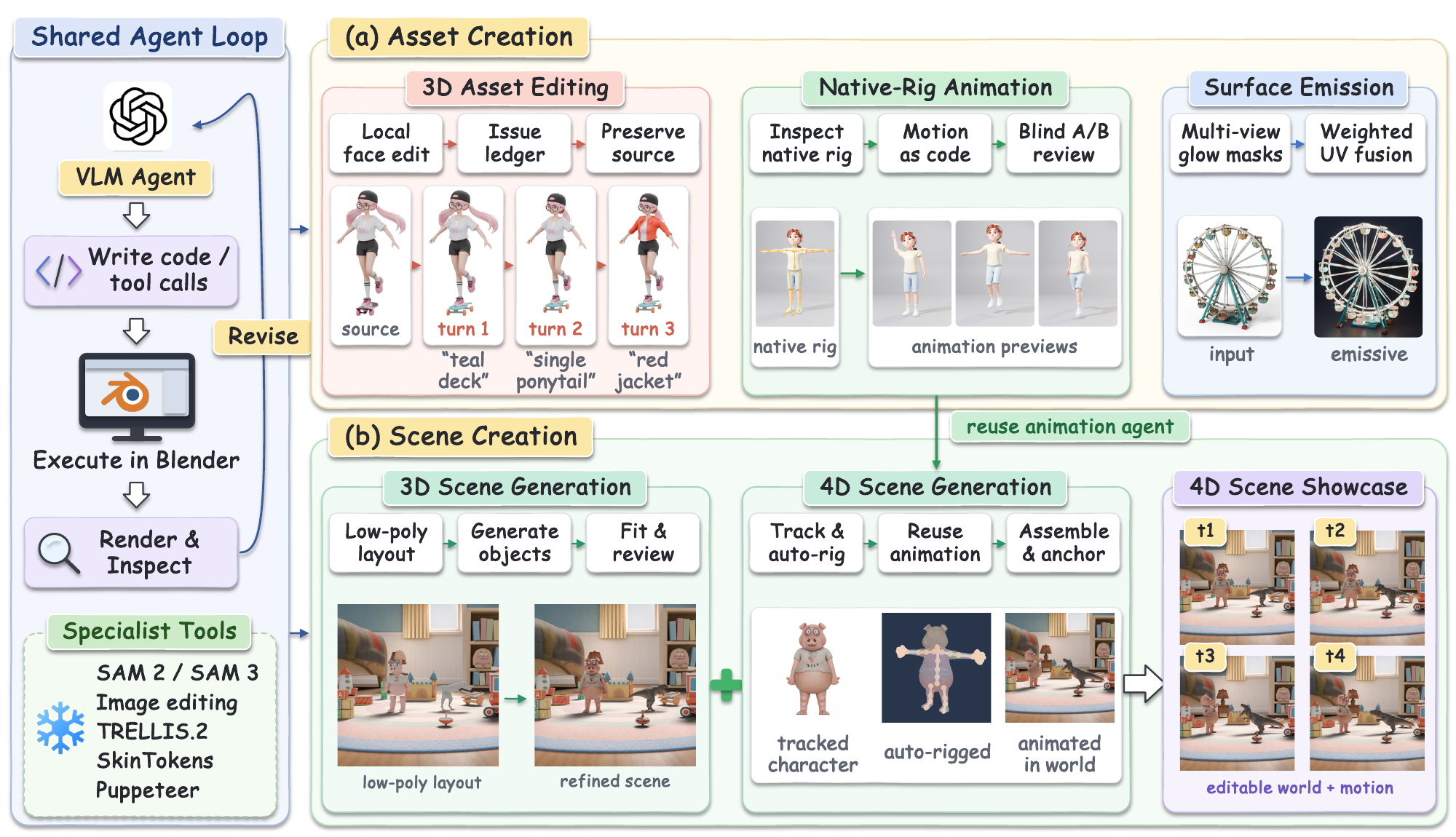}
  \vspace{-20pt}
  \caption{\textbf{Overview of our unified agent framework for 3D and 4D content creation.}
\textbf{(a) Asset creation.} 3D asset editing edits only selected faces and
preserves the source mesh's remaining faces, vertices, UVs, and materials; native-rig animation writes keyframed bone transforms; surface emission segments glow per view and fuses masks on the texture atlas.
\textbf{(b) Scene creation.} 3D scene generation fixes low-poly layouts, completes references and fits generated meshes; 4D scene generation reuses animation agent to track, auto-rig and anchor characters.}
 \vspace{-16pt}
  \label{fig:overview}
\end{figure*}

All five tasks share one loop: A vision-language LLM agent writes Blender Python, runs it in a workspace that only contains task inputs, renders the result and revises the asset until renders match the goal without task-specific training. Specialist models for segmentation, image editing, 3D generation and rigging are tool calls that write files into the same workspace. 
It never sees the ground-truth geometry or evaluation scores. The harness loop is composed of four parts: in editing it keeps an issue ledger and a record of unchangeable areas; in animation it exposes the native rig and renders paired previews; in emission it fuses per-view masks on the texture atlas; in scene generation it completes references and fits detailed meshes into low-poly layout. The agent decides what to change, how to change, and when the result is good enough to stop and deliver.

\subsection{3D asset editing}
\label{sec:method_3d_edit}

The unified agent supports single-turn and multi-turn editing, and each turn may add, remove, replace or retexture.
At turn $t$ the input is the previous output $\mathcal{M}_{t-1}$, an instruction $I_t$, a conditioning image $C_t$ and
a fixed camera $K$; following EditHero~\cite{edithero}, each turn edits the previous output. The agent selects the
target faces with PartField~\cite{partfield}, multi-view face voting or SAM~\cite{sam2,sam3} and edits them in Blender
Python. Removal and replacement delete the selected faces and add new faces to repair the surface, while all other
vertices, connectivity, UVs and materials stay as they are. Before editing, the agent declares which objects it may
change or add. All other objects are protected, and the harness marks the delivery incomplete if any protected vertex
differs from the source. Projection and contact measurements tell the agent whether an error comes from shape or from
placement. The agent renders the current asset and records what still differs from the instruction in an issue ledger, and then
works on the most serious open issue. Each issue is linked to the renders that show it, so once the asset changes,
earlier renders no longer count as evidence. 

\subsection{Rig-based animation}
\label{sec:method_4d_asset}

Given a rigged asset in rest pose (vertices $\{\mathbf{x}_i\}$, $J$ bones, skinning weights $\{\omega_{ij}\}$) and a
video $\{I_t\}_{t=1}^{T}$, we predict per-frame bone poses $\boldsymbol{\theta}_t$ and an armature transform
$\mathbf{G}_t$ so that linear blend skinning,
\begin{equation}
  \mathbf{x}_i(t)=\mathbf{G}_t\sum_{j=1}^{J}\omega_{ij}\,\mathbf{B}_j(\boldsymbol{\theta}_t)\,\mathbf{x}_i,
  \label{eq:lbs}
\end{equation}
reproduces the motion in the video. The mesh, rest skeleton and weights stay unchanged. Each bone is described by its head and tail, parent, local frame and the surface region it moves. The agent works out from this geometry which chains are legs, neck or tail. It then writes the whole animation in Blender Python as keyframed bone and object transforms, without a motion prior or pose solver. A preview
tool shows each reference frame next to the animated frame with the same index.

The agent keeps every candidate unchanged once saved and compares at least four. Each comparison goes to an
independent agentic reviewer that sees only the video and two candidates, shown as A and B in random order, and comments on
action phase, facing direction, large joint motions and contacts. The agent decides which advice to follow and which
candidate to deliver. The harness exports a GLB and checks every frame both in Blender and in the exported file.

\subsection{Surface emission}
\label{sec:method_emission}

Given a source mesh $\mathcal{M}$, its non-emissive base color $\mathbf{a}$ and a thumbnail, we predict an emission map $\mathbf{E}$ on texture atlas $\mathcal{U}$: a binary selection $s(t)\in\{0,1\}$ of the texels $t$ that emit, and their color. The harness unwraps one non-overlapping atlas and renders $V{=}4$ base-color views around the object. Ray casting
records the first texel hit at every pixel, so occluded surfaces receive no projection. In each view the agent segments
the emissive parts with SAM~\cite{sam2}, SAM~3~\cite{sam3} or CIE76 color thresholding ($\tau_c{=}15$),
checks a zoomed view of the boundary, and refines the prompt or adds an exclusion mask until it accepts the mask $m_v$.
Surfaces that no view sees get extra cameras, up to 12 views in total.

The harness fuses the masks on the atlas, weighting each view by how reliably it observes a texel:
\begin{equation}
  \begin{aligned}
  w(v,t) &= \overline{\left|\mathbf{n}\cdot\mathbf{d}_v\right|^2}\;\min\!\left(\frac{N(v,t)}{4},\,1\right),\\
  P(t)&=\textstyle\sum_v w(v,t)\,m_v(t),\\
  Q(t)&=\textstyle\sum_v w(v,t)\,\bigl(1-m_v(t)\bigr),\\
  s(t)&=\mathbf{1}\bigl[P(t)>Q(t)\bigr].
  \end{aligned}
  \label{eq:emission_fusion}
\end{equation}
The first factor of $w$ is the mean squared cosine between surface normal and viewing direction over the pixels that
project to $t$, and $N(v,t)$ counts the visible pixels whose $2{\times}2$ footprint covers $t$. A view that does not see
$t$ has zero weight, so it cannot vote against a selection made in other views. A selected texel emits its own base color and the rest emit nothing, $\mathbf{E}(t)=s(t)\mathbf{a}(t)$ in linear space, so a lamp with parts of different colors keeps each of them.

\subsection{3D scene generation}
\label{sec:method_3d_scene}

Given an image $I$ and its camera, we generate $N$ objects $\{(\mathcal{M}_n,\mathbf{T}_n)\}_{n=1}^{N}$ standing on the
ground plane $z=0$. The agent fixes the layout first. It edits a Blender script that starts from a seed scene with the source camera, the ground and an empty \texttt{Objects} node, and builds each object under that node as a low-poly proxy from
primitives. It places ground contacts by intersecting the camera ray through a pixel with $z=0$, renders from the source
camera, and adjusts position, size and orientation until the render matches the input. Following stages only change the
geometry, not the layout.

For each object, the agent crops the input image around the visible part of the proxy silhouette (12\% margin, $512^2$) and fills occluded pixels with a render of the proxy. gpt-image-2 completes this crop into a pixel-aligned reference, accepted only if its BiRefNet~\cite{birefnet} mask has IoU of at least $0.55$ with the original crop. TRELLIS.2~\cite{trellis2} generates mesh $\hat{\mathcal{M}}_n$ from the reference, and harness fits it to proxy with a uniform scale, a rotation $\mathbf{R}_z(\psi)$ about vertical axis and a translation minimizing silhouette difference in source camera. The mesh inherits proxy's placement and stays upright. A reviewer keeps either generated mesh or the proxy for each object.

\subsection{4D scene generation}
\label{sec:method_4d_scene}

Given a monocular video and its camera, we reconstruct the first frame with \cref{sec:method_3d_scene} and animate the
moving character with \cref{sec:method_4d_asset}. To find the character, we render the visible region $\Omega_n$ of
each reconstructed object in the first frame. We measure temporal change at pixel $p$ as
\begin{equation}
  D(p)=\frac{1}{T}\sum_{t=1}^{T}\bigl\|I_t(p)-I_1(p)\bigr\|_1,
  \label{eq:motion_score}
\end{equation}
and the score $S_n$ of object $n$ sums $D$ over a $15{\times}15$ dilation of $\Omega_n$, which tolerates small
misalignment of the reconstruction. The object with the largest $S_n$ is the character. We prompt SAM~2~\cite{sam2}
with its projected region, take the largest candidate mask, subtract the masks of objects in front of it, and track the
character and these occluders together.
The character is composited on white and cropped by one box that covers it in all frames, which
gives an object-centered reference video and keeps its trajectory.

A transform $\mathbf{C}$ makes the extracted character upright, turns it so that the camera's horizontal viewing
direction is $+y$, and moves the center of its base to the origin. SkinTokens~\cite{skintokens} predicts the skeleton
and Puppeteer~\cite{puppeteer} the skinning weights. After animation, $\mathbf{C}^{-1}$ maps the result back into the
scene, where it replaces the static object. Because the agent writes root motion without metric scale, the first
animated frame can drift from the reconstruction. We therefore align the posed first frame to the rest shape through a similarity transform process, estimated with the method proposed by Umeyama~\cite{umeyama},
\begin{equation}
  (\sigma,\mathbf{Q},\mathbf{b})=\mathop{\mathrm{arg\,min}}_{\sigma,\mathbf{Q},\mathbf{b}}\sum_i
  \bigl\|\sigma\mathbf{Q}\,\mathbf{x}_i(1)+\mathbf{b}-\mathbf{x}_i\bigr\|^2,
  \label{eq:anchor}
\end{equation}
and apply it to all frames, so the articulated motion is preserved relative to the anchored first frame. If tracking or rigging fails, the static reconstruction is kept.

%% file: sec/4_experiments.tex
\newcommand{\ours}{Ours}
\newcommand{\benchthree}{CompoScene-3D}
\newcommand{\benchfour}{CompoScene-4D}

\section{Experiments}
\label{sec:exp}
\setlength{\textfloatsep}{8pt plus 2pt minus 2pt}\setlength{\dbltextfloatsep}{8pt plus 2pt minus 2pt}
\setlength{\floatsep}{6pt plus 2pt minus 2pt}\setlength{\dblfloatsep}{6pt plus 2pt minus 2pt}
\setlength{\abovecaptionskip}{4pt}\setlength{\belowcaptionskip}{0pt}
\makeatletter\setlength{\@fptop}{0pt}\setlength{\@fpsep}{10pt plus 2pt}\setlength{\@fpbot}{0pt plus 1fil}
\setlength{\@dblfptop}{0pt}\setlength{\@dblfpsep}{10pt plus 2pt}\setlength{\@dblfpbot}{0pt plus 1fil}\makeatother

We evaluate our framework on five generation tasks that span assets and scenes as well as static and dynamic content:
3D asset editing, emission generation, 4D asset animation, single-image 3D scene generation, and video-to-4D scene
generation. The first three use established benchmarks; for the two scene tasks, we construct \benchthree{} and
\benchfour{}.

\renewcommand{\dbltopfraction}{0.85}\renewcommand{\topfraction}{0.85}\renewcommand{\dblfloatpagefraction}{0.9}\renewcommand{\floatpagefraction}{0.9}

\begin{table*}[t]
\centering
\caption{Quantitative comparison on EditHero (457 chains, 2{,}755 edits, self-rollout). $E$: edit region, $U$: unedited
region, $W$: whole object. CD is $\times100$. Best in bold, second best underlined.}
\label{tab:edithero}
\setlength{\tabcolsep}{4pt}
\small
\begin{tabular}{lccccccccccc}
\toprule
\multirow{2}{*}{Method}
 & \multirow{2}{*}{IF $\uparrow$}
 & \multicolumn{2}{c}{Consistency $\uparrow$}
 & \multicolumn{3}{c}{Geometry ($W$)}
 & \multicolumn{3}{c}{LPIPS $\downarrow$}
 & \multirow{2}{*}{PSNR$_W$ $\uparrow$}
 & \multirow{2}{*}{Drift$_U$ $\downarrow$} \\
\cmidrule(lr){3-4} \cmidrule(lr){5-7} \cmidrule(lr){8-10}
 & & CC$^{\text{prev}}$ & CC$^{0}$
 & F $\uparrow$ & IoU $\uparrow$ & CD $\downarrow$
 & $E$ & $U$ & $W$ & & \\
\midrule
3DEditFormer~\cite{threeditformer}
 & 0.191 & 0.783 & 0.487 & 0.523 & 0.411 & 3.04
 & 0.529 & 0.200 & 0.215 & 17.8 & 0.0157 \\
VoxHammer~\cite{voxhammer}
 & 0.32 & 0.71 & 0.453 & 0.52 & 0.380 & 2.69
 & 0.521 & 0.205 & 0.219 & 15.2 & 0.022 \\
PartFlow~\cite{partflow}
 & 0.118 & 0.878 & 0.688 & 0.647 & 0.519 & 2.28
 & 0.529 & 0.174 & 0.190 & 18.1 & 0.0128 \\
Nano3D~\cite{nano3d}
 & 0.309 & 0.890 & 0.728 & 0.696 & 0.581 & 2.09
 & 0.487 & 0.154 & 0.170 & 18.9 & 0.0076 \\
Alchemy3D~\cite{alchemy3d}
 & 0.410 & 0.680 & 0.346 & 0.446 & 0.348 & 4.44
 & 0.516 & 0.218 & 0.233 & 17.6 & 0.0177 \\
Codex agent (GPT-6 Astra)
 & \underline{0.619} & \underline{0.981} & \underline{0.920}
 & \underline{0.832} & \underline{0.806} & \underline{0.80}
 & \underline{0.222} & \underline{0.054} & \underline{0.062}
 & \underline{26.9} & \underline{0.0018} \\
\midrule
\textbf{\ours}
 & \textbf{0.638} & \textbf{0.982} & \textbf{0.922}
 & \textbf{0.840} & \textbf{0.814} & \textbf{0.79}
 & \textbf{0.193} & \textbf{0.048} & \textbf{0.056}
 & \textbf{27.7} & \textbf{0.0017} \\
\bottomrule
\end{tabular}
\end{table*}

\begin{table}[t]
\centering
\caption{Quantitative comparison on the LightGenBench test split (200 shapes), official evaluator. Other methods as
reported by LightGenBench. Best in bold, second best underlined.}
\label{tab:lightgen}
\begin{tabular}{lccc}
\toprule
Method & IoU $\uparrow$ & MAE $\downarrow$ & PSNR $\uparrow$ \\
\midrule
TEXGen-Emission     & 0.092 & 0.025 & 25.10 \\
Hunyuan3D-Emission  & 0.117 & \underline{0.022} & 26.46 \\
SegviGen-Emission   & 0.147 & 0.041 & 24.50 \\
TRELLIS.2-Emission  & \underline{0.271} & \textbf{0.021} & \underline{26.81} \\
\midrule
\textbf{\ours}      & \textbf{0.338} & 0.027 & \textbf{27.23} \\
\bottomrule
\end{tabular}
\end{table}

\begin{table}[t]   
\centering
\scriptsize
\setlength{\tabcolsep}{2pt}
\caption{Quantitative comparison on Motion-80. All methods receive the rest mesh, original skeleton and skinning weights,
and the video. SR is computed over the full split (64 short, 16 long); quality metrics are averaged over the subjects that
all methods complete (61 short, 15 long), with appearance over three held-out views. Averaged over the short and long
sequences (weighted by subjects). $^{\dagger}$Uses a 512-joint input
limit. Best in bold, second best underlined.}
\label{tab:motion80}
\resizebox{\linewidth}{!}{%
\begin{tabular}{lrrrrrrr}
\toprule
Method & SR (\%) $\uparrow$ & CD $\downarrow$ & F$_{0.02}$ $\uparrow$ & LPIPS $\downarrow$ & CLIP $\uparrow$ & FVD $\downarrow$ & DreamSim $\downarrow$ \\
\midrule
CAST$^{\dagger}$~\cite{castmocap} & 100 & 0.2679 & 0.1792 & 0.1816 & 0.8651 & 1842.35 & 0.2008 \\
TopoCap~\cite{topocap} & 95 & 1.3051 & 0.1046 & 0.2063 & 0.8135 & 2337.03 & 0.2616 \\
MorphGS~\cite{morphgs} & 100 & \underline{0.2080} & \underline{0.2177} & \underline{0.1498} & \underline{0.8842} & \underline{1562.99} & \underline{0.1672} \\
MoCapAnything V2$^{\dagger}$~\cite{mocapv2} & 100 & 0.7110 & 0.0967 & 0.2141 & 0.7529 & 2530.28 & 0.3563 \\
\midrule
\textbf{\ours} & 100 & \textbf{0.1179} & \textbf{0.3686} & \textbf{0.1214} & \textbf{0.9451} & \textbf{898.77} & \textbf{0.0883} \\
\bottomrule
\end{tabular}}
\end{table}

\begin{figure*}[t]
\centering
\begin{minipage}[t]{0.49\linewidth}
\centering
\includegraphics[width=\linewidth]{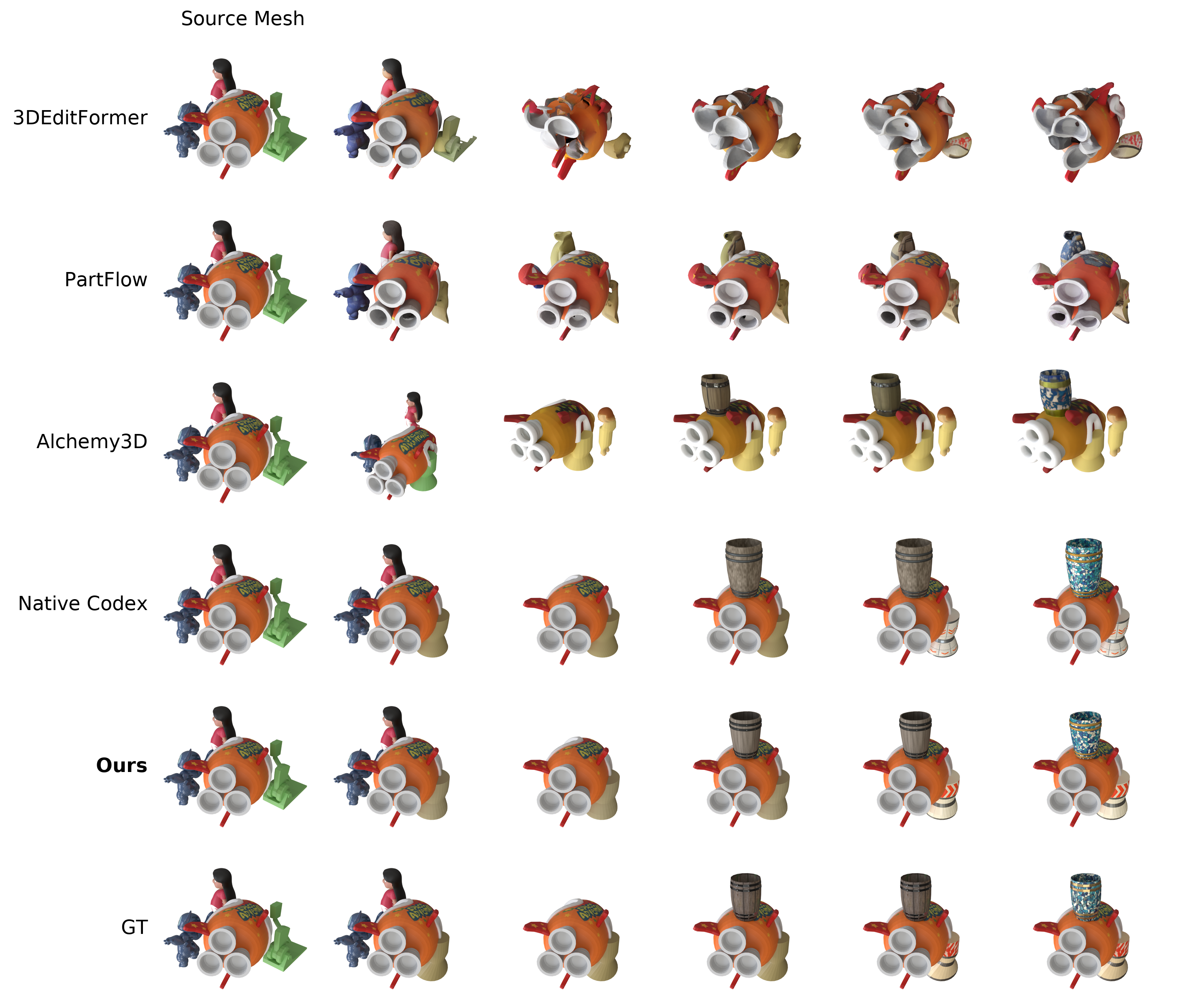}
\caption{Qualitative comparison on one EditHero chain (self-rollout): the source mesh and the results after each turn.}
\label{fig:edithero}
\end{minipage}\hfill
\begin{minipage}[t]{0.49\linewidth}
\centering
\includegraphics[width=\linewidth]{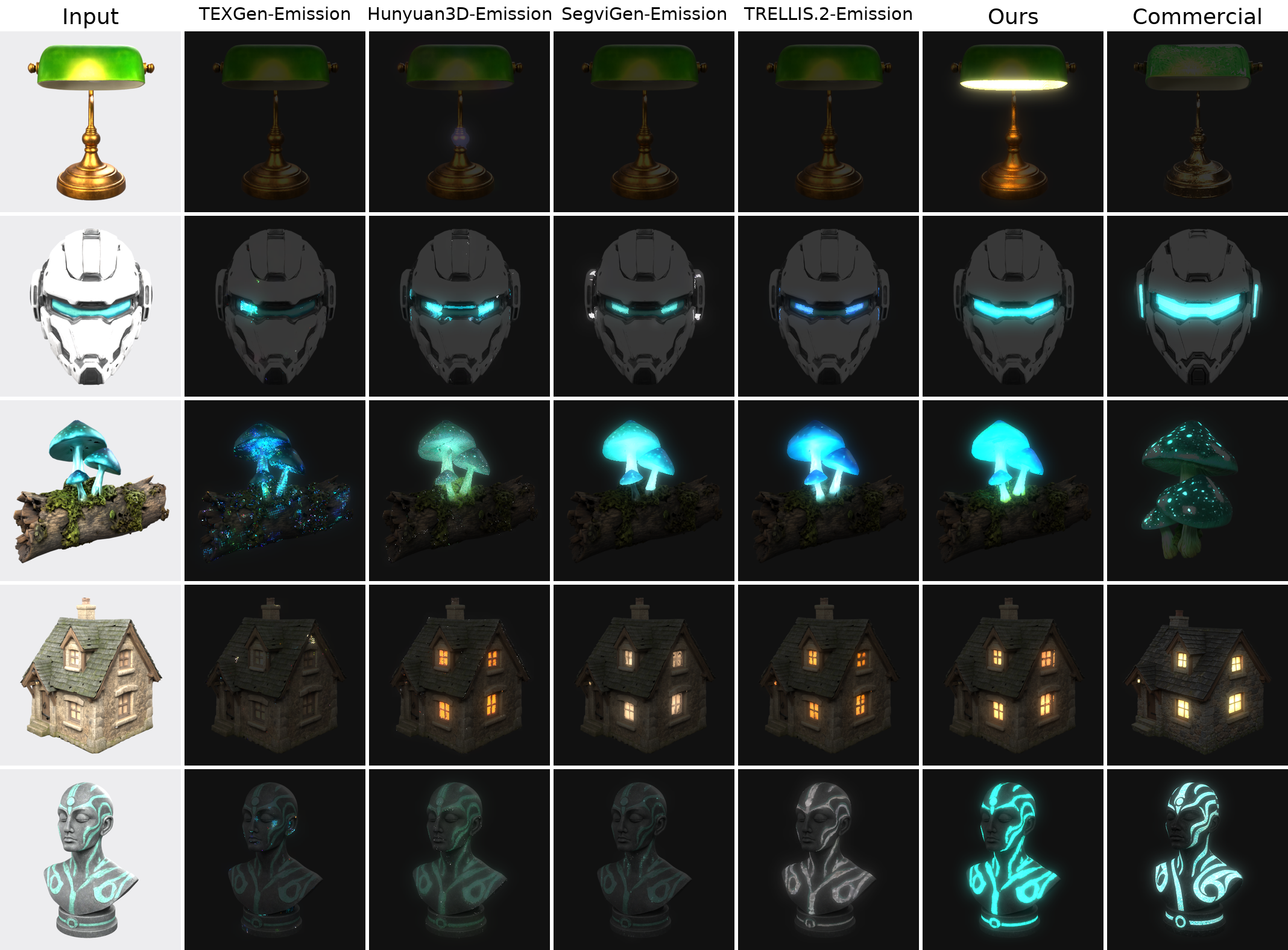}
\caption{Emission generation on our own assets outside the dataset, which have no ground truth; last column shows a commercial model.}
\label{fig:emission_own}
\end{minipage}
\end{figure*}

\begin{table}[t]   
\centering
\caption{Quantitative comparison on \benchthree{} (120 scenes, single-image input). Succ.\ is the number of scenes a
method produced; failed scenes count as zero IoU, and PL and N-CLIP are averaged over produced scenes. $^\ast$Mean of 3
runs. Best in bold, second best underlined.}
\label{tab:scene3d}
\setlength{\tabcolsep}{3pt}
\resizebox{\linewidth}{!}{%
\begin{tabular}{lccccccc}
\toprule
\multirow{2}{*}{Method} & \multirow{2}{*}{Succ.} & \multicolumn{2}{c}{Scene-IoU $\uparrow$} & \multicolumn{2}{c}{Object-IoU $\uparrow$} & \multirow{2}{*}{PL $\downarrow$} & \multirow{2}{*}{N-CLIP $\downarrow$} \\
\cmidrule(lr){3-4} \cmidrule(lr){5-6}
 & & AABB & OBB & AABB & OBB & & \\
\midrule
MIDI-3D~\cite{midi}              & 120 & 0.660 & 0.458 & 0.187 & 0.154 & 1.99 & 1.23 \\
SceneGen~\cite{scenegen}         & 120 & 0.557 & 0.426 & 0.114 & 0.094 & 3.52 & 1.56 \\
SceneMaker~\cite{scenemaker}     & 120 & 0.569 & 0.379 & 0.112 & 0.092 & 3.42 & 1.63 \\
Mira-Scene~\cite{mirascene}      & 119 & 0.669 & 0.449 & 0.207 & 0.166 & 2.37 & 1.44 \\
OVOW~\cite{ovow}                 & 113 & 0.544 & 0.390 & 0.151 & 0.122 & 2.29 & 1.33 \\
VIGA~\cite{viga}                 & 120 & 0.643 & 0.434 & 0.134 & 0.101 & 2.55 & 1.54 \\
GPT-6 Astra direct$^\ast$        & 120 & \underline{0.801} & \underline{0.565} & \underline{0.399} & \underline{0.329} & \textbf{1.21} & \underline{1.18} \\
\midrule
\textbf{\ours}                   & 120 & \textbf{0.836} & \textbf{0.589} & \textbf{0.459} & \textbf{0.372} & \underline{1.30} & \textbf{0.83} \\
\bottomrule
\end{tabular}}
\end{table}

\begin{table*}[t]
\centering
\caption{Quantitative comparison on \benchfour{} (120 scenes, video input). Scene metrics are taken at the first, middle and
last frames; character metrics are averaged over all frames. The single-image methods are run independently on each of the
three frames and scored against the ground truth at that frame; they output no animation, so the character metrics do not
apply. Failed scenes (frames) count as zero IoU and zero F-score; PL, N-CLIP and character CD are averaged over produced
scenes. Best in bold, second best underlined.}
\label{tab:scene4d}
\footnotesize
\setlength{\tabcolsep}{3pt}   
\begin{tabular}{lcccccccccc}
\toprule
\multirow{3}{*}{Method} & \multicolumn{6}{c}{Scene} & \multicolumn{4}{c}{Deformable character} \\
\cmidrule(lr){2-7} \cmidrule(lr){8-11}
 & \multicolumn{2}{c}{Scene-IoU $\uparrow$} & \multicolumn{2}{c}{Object-IoU $\uparrow$} & \multirow{2}{*}{PL $\downarrow$} & \multirow{2}{*}{N-CLIP $\downarrow$} & \multicolumn{2}{c}{Local} & \multicolumn{2}{c}{World} \\
\cmidrule(lr){2-3} \cmidrule(lr){4-5} \cmidrule(lr){8-9} \cmidrule(lr){10-11}
 & AABB & OBB & AABB & OBB & & & CD $\downarrow$ & F@0.02 $\uparrow$ & CD $\downarrow$ & F@5\% $\uparrow$ \\
\midrule
\multicolumn{11}{l}{\textit{Single image, per frame}} \\
MIDI-3D~\cite{midi}       & \underline{0.667} & \underline{0.454} & 0.151 & \underline{0.129} & \underline{1.81} & \underline{1.14} & -- & -- & -- & -- \\
SceneGen~\cite{scenegen}  & 0.597 & 0.416 & 0.123 & 0.101 & 3.09 & 1.44 & -- & -- & -- & -- \\
SceneMaker~\cite{scenemaker} & 0.597 & 0.379 & 0.119 & 0.099 & 3.00 & 1.51 & -- & -- & -- & -- \\
Mira-Scene~\cite{mirascene} & 0.618 & 0.410 & \underline{0.152} & 0.125 & 2.59 & 1.66 & -- & -- & -- & -- \\
VIGA~\cite{viga}          & 0.614 & 0.357 & 0.067 & 0.049 & 2.50 & 1.90 & -- & -- & -- & -- \\
\midrule
\multicolumn{11}{l}{\textit{Video}} \\
OVOW~\cite{ovow} & 0.446 & 0.327 & 0.115 & 0.098 & 2.90 & 1.53 & \underline{0.491} & \underline{0.093} & \underline{0.627} & \underline{0.162} \\
\midrule
\textbf{\ours}   & \textbf{0.776} & \textbf{0.546} & \textbf{0.340} & \textbf{0.285} & \textbf{1.33} & \textbf{0.87} & \textbf{0.202} & \textbf{0.216} & \textbf{0.227} & \textbf{0.450} \\
\bottomrule
\end{tabular}
\end{table*}

\subsection{Implementation Details}
\label{sec:exp_impl}

All tasks run the same agent framework without task-specific training: a vision-language model plans, writes Blender
programs, inspects renders of the result and revises it, while foundation models serve as tools. We use GPT-6 Astra as
the agent for asset editing (high reasoning effort), scene reconstruction and animation (extra-high reasoning effort), and
GPT-6.1 Sol for emission generation. Tools include SAM~2~\cite{sam2} and SAM~3~\cite{sam3} for segmentation and
tracking, TRELLIS.2~\cite{trellis2} for object meshes, and SkinTokens~\cite{skintokens} and Puppeteer~\cite{puppeteer}
for automatic rigging. For scenes, the agent first lays out the scene with Blender primitives, then completes each
object's reference image, generates its mesh and fits it to the layout. The agent never sees ground-truth geometry,
motion or scores. Emission generation uses at most 30 model requests per shape.

\subsection{Benchmark Setup}
\label{sec:exp_setup}

\paragraph{Evaluation datasets.}
\emph{EditHero}~\cite{edithero} contains 457 multi-turn editing chains with 2{,}755 part-level edits (add, remove,
replace, retexture) over 252 objects and up to 30 turns per chain; every turn starts from the method's own previous output
(self-rollout). \emph{LightGenBench}~\cite{lightgenbench} asks for the RGB surface emission of a shape from its geometry,
its non-emissive PBR material and one thumbnail; we use its 200 test shapes. \emph{Motion-80}~\cite{motion3to4} contains
64 short and 16 long animated characters, each with a monocular video; every method receives the rest mesh with its
original skeleton and skinning weights (all animation removed) and the video.
For scenes, we render two synthetic benchmarks in Blender. \emph{\benchthree{}} contains 120 static scenes, each composing
3--5 objects from Objaverse-OA~\cite{objaverseoa} (461 objects over 362 LVIS categories) at the typical real-world size of
their category on a ground plane, without overlap, under one of 30 Poly Haven HDRIs; the input is one $512^2$ view in which
every object is visible. \emph{\benchfour{}} extends each \benchthree{} scene with an animated Motion-80 character, so that
every scene contains a deformable object next to 2--4 static ones; a fixed camera records the scene as a video of up to
108 frames at 12 FPS. Both benchmarks provide ground-truth instance meshes and cameras, and \benchfour{} also provides the
per-frame vertices of the character.

\paragraph{Baselines.}
For editing, we compare against the non-agentic editors 3DEditFormer~\cite{threeditformer}, VoxHammer~\cite{voxhammer}, PartFlow~\cite{partflow},
Nano3D~\cite{nano3d}, Alchemy3D~\cite{alchemy3d}, and a Codex agent driven by the same model as ours. For emission, we compare
against the methods reported by LightGenBench. For animation, we compare against CAST~\cite{castmocap},
TopoCap~\cite{topocap}, MorphGS~\cite{morphgs} and MoCapAnything V2~\cite{mocapv2}, all given the same rest mesh, rig and
video. For 3D scenes, we compare against the compositional generators MIDI-3D~\cite{midi}, SceneGen~\cite{scenegen},
SceneMaker~\cite{scenemaker} and Mira-Scene~\cite{mirascene}, the agentic pipelines OVOW~\cite{ovow} and VIGA~\cite{viga},
and a direct baseline where GPT-6 Astra writes the whole Blender scene from the image. All methods receive the same RGB image. Methods that accept camera parameters also receive the same ground-truth intrinsics and extrinsics, which are used for rendering and evaluation across all methods. MIDI-3D, SceneGen and SceneMaker use the same automatic instance masks, while Mira-Scene runs its complete
default pipeline with its own VLM-guided segmentation, using GPT-6.1 Sol as the VLM. For 4D scenes, we compare against OVOW, which like ours reconstructs instance-level 4D scenes from a video, and
run MIDI-3D, SceneGen, SceneMaker, Mira-Scene and VIGA independently on the first, middle and last frame of each video,
the frames at which the scene metrics are taken, with the same settings as for 3D scenes.

\begin{figure*}[t]
\centering
\includegraphics[width=\linewidth]{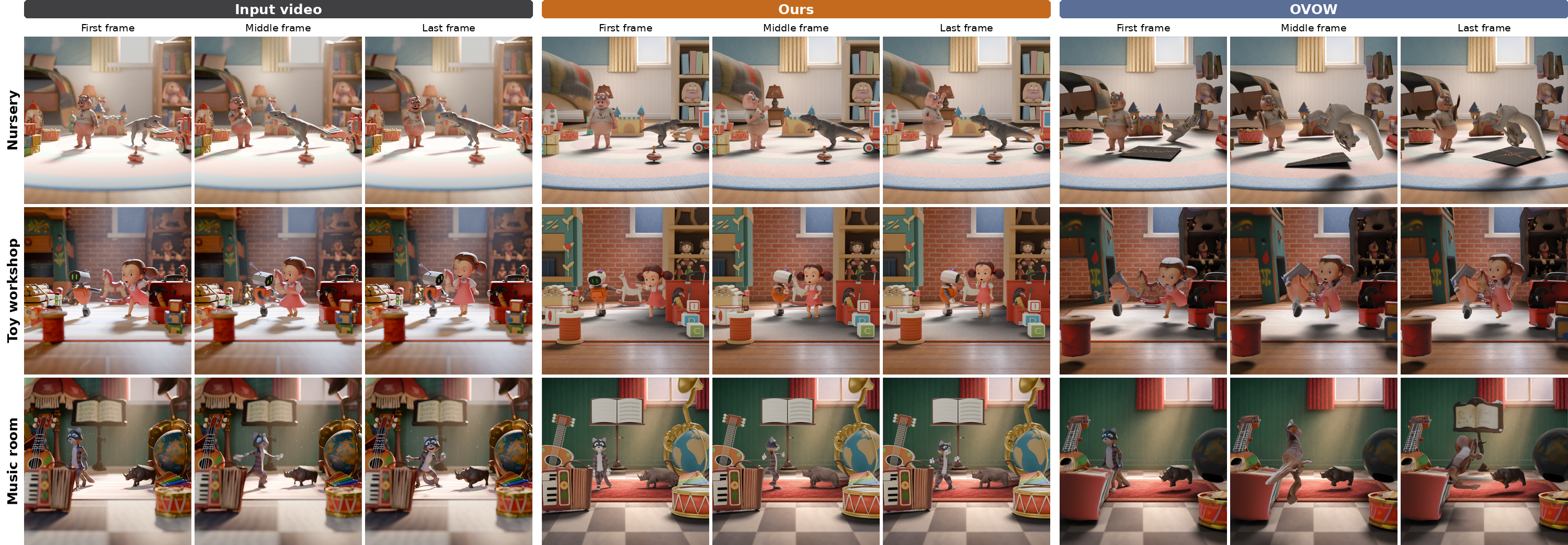}
\caption{Video-to-4D comparison with OVOW on three animated-film videos outside the benchmark, at their first, middle and
last frames: input video, our reconstructed instance-level 4D scene, and OVOW, both rendered from the input camera with
their reconstructed backgrounds.}
\label{fig:demos}
\end{figure*}

\paragraph{Metrics.}
For editing we follow EditHero: instruction following inside the edit region (IF), content consistency outside it with
respect to the previous turn and the initial object (CC$^{\text{prev}}$, CC$^{0}$), whole-object F-score, IoU and Chamfer
distance, LPIPS~\cite{add_lpips} on the edited, unedited and whole renders, PSNR and image drift. For emission we use the official IoU of
the emission masks, MAE and PSNR. For animation we report the success rate (SR), Chamfer distance and F-score at 0.02 after
first-frame alignment, and LPIPS, CLIP~\cite{add_clip} similarity, FVD~\cite{add_fvd} and DreamSim~\cite{add_dreamsim} on three held-out views. For scenes we use the
OVOW-Bench evaluator~\cite{ovow}: Scene-IoU of the union of object boxes (AABB, OBB), Object-IoU after Hungarian matching
with a penalty for missing or extra objects, photometric loss (PL, $100\times$ normalized RGB MSE) and N-CLIP
($10\times(1-\text{CLIP similarity})$), computed at the first, middle and last frames for \benchfour{}. For PL and
N-CLIP every prediction is rendered under the ground-truth environment light and ground plane, as the ground-truth scene
is, so that both measure the reconstructed objects rather than the background. For the
deformable character in \benchfour{} we additionally report per-frame Chamfer distance and F-score, both after aligning
the first frame (local) and in the aligned scene space (world). A scene that a method fails to produce counts as zero.

\subsection{Quantitative Comparison}
\label{sec:exp_quant}

\Cref{tab:edithero,tab:lightgen,tab:motion80,tab:scene3d,tab:scene4d} report the comparison on the
five tasks; in each table the best and second-best results are highlighted.

\paragraph{3D asset editing.}
On EditHero, \ours{} obtains the best score on every single metric (\cref{tab:edithero}). Against the non-agentic editors, it more than triples instruction following (0.638 vs.\ 0.191 for
3DEditFormer), raises whole-object IoU from 0.519 to 0.814, and keeps almost all unedited content
(CC$^{\text{prev}}$ 0.982). Against the Codex agent that uses the same model, \ours{} improves instruction following and
every appearance metric, most clearly in the edited region (LPIPS$_E$ 0.193 vs.\ 0.222). The lead holds for every edit
type, and is largest for retexturing and replacement.

\paragraph{Emission generation.}
On LightGenBench, \ours{} raises emission IoU from 0.271 to 0.338 and PSNR from 26.81 to 27.23 over the strongest method (\cref{tab:lightgen}). Against our run of the TRELLIS.2-Emission checkpoint on the same shapes, most of the gain comes from precision (0.511 vs.\ 0.355): \ours{} lights far fewer surfaces that should stay dark.
\begin{figure}[t]
  \centering
  \includegraphics[width=\columnwidth]{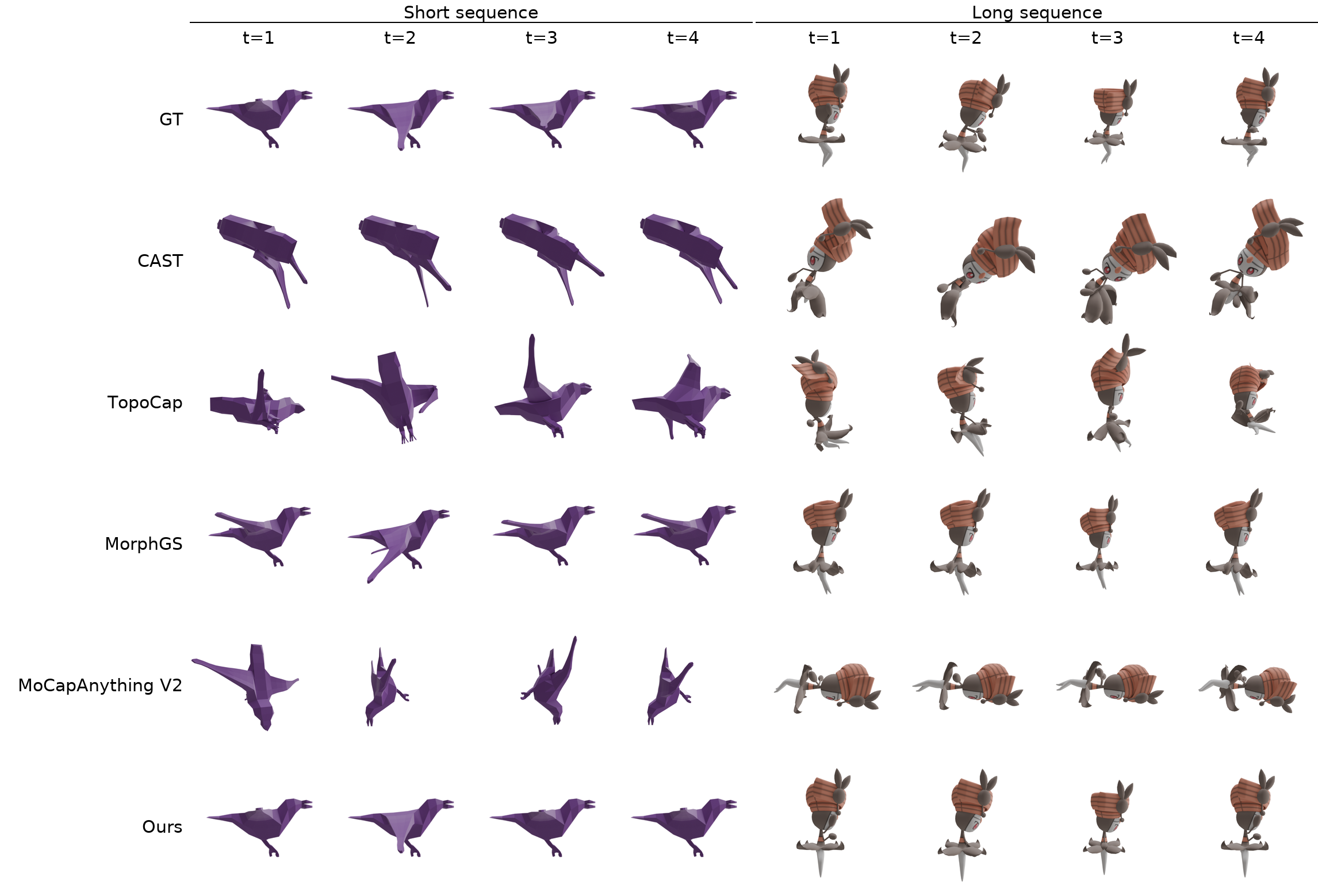}
  \caption{Qualitative comparison on Motion-80 for a short
  sequence (left) and a long sequence (right), each shown
  at four time steps. All methods receive the same rest
  mesh and native rig; GT denotes the ground-truth animation.}
  \label{fig:motion80}
\end{figure}
\paragraph{4D asset animation.}
On Motion-80, \ours{} leads every quality metric while animating every subject (\cref{tab:motion80}). Averaged over the
short and long sequences, it reduces CD from 0.208 to 0.118 and raises F-score from 0.218 to 0.369 relative to the
strongest alternative, MorphGS, and lowers FVD from 1563 to 899; the lead holds on both splits (short: CD 0.115 vs.\
0.209, long: CD 0.132 vs.\ 0.203).

\paragraph{3D scene generation.}
On \benchthree{}, \ours{} obtains the best Scene-IoU, Object-IoU and N-CLIP (\cref{tab:scene3d}). Its Object-IoU (AABB
0.459) is more than twice that of the best compositional generator (0.207, Mira-Scene) and three times that of the agentic
pipelines OVOW (0.151) and VIGA (0.134), and it remains above MIDI-3D given ground-truth instance masks (0.254). It also
improves on the direct GPT-6 Astra baseline (0.399), which uses the same model without our framework.

\paragraph{4D scene generation.}
On \benchfour{}, \ours{} outperforms OVOW on every metric, with Object-IoU (AABB) 0.340 against 0.115
(\cref{tab:scene4d}). On the deformable character it reduces the
world-space Chamfer distance by a factor of 2.8 (0.227 vs.\ 0.627) and nearly triples the world F-score (0.450 vs.\ 0.162).
The single-image methods reconstruct each frame on its own and therefore see the character in its current pose at every
scored frame; even so, \ours{} leads all five of them on every scene metric while producing one consistent, animated
scene, with Object-IoU (AABB) 0.340 against 0.152 for the strongest of them, Mira-Scene, and Scene-IoU (AABB) 0.776
against 0.667 for MIDI-3D.

\paragraph{Where the gains come from.}
The gains share a common source: \ours{} expresses its output as an explicit Blender program and checks the rendered
result before delivering it. In editing, a program changes only the parts named by the instruction, which keeps the
unedited content intact across long chains, whereas the non-agentic editors regenerate the whole object at every turn.
In scenes, the agent first fixes the layout with primitives and then reconstructs and fits each object to it, so every
object keeps its own identity, position and scale; this is reflected in Object-IoU, where the gap to the baselines is
largest. In animation, the agent drives the given skeleton and compares candidate motions against the video, so the
character keeps its shape while it moves. The baselines that use the same model, the Codex agent in editing and GPT-6
Astra direct in 3D scene generation, trail \ours{} on almost every metric, indicating that these gains come from the
framework rather than the underlying model.

\subsection{Qualitative Comparison}
\label{sec:exp_qual}

In \cref{fig:edithero}, the non-agentic editors distort the object from the first turns on and drift further with every turn, and the Codex agent encounters size issues; \ours{} applies each edit and keeps the rest
of the object.
On our own assets (\cref{fig:emission_own}), which lie outside the training data of every baseline, the baselines light almost nothing or scattered points, while \ours{} lights the light strips of the armor and the carved patterns of the sculpture.
\Cref{fig:demos} compares \ours{} with OVOW on three animated-film videos outside the benchmark, a nursery, a toy workshop and a music room, each with two characters among many props. \ours{} recovers the furniture and props of each set as
separate instances with their layout, reconstructs the room behind them, and animates both characters throughout the video. OVOW distorts the characters and their poses (\eg, the dinosaur, the robot and the cat) and loses several props.

\FloatBarrier  

%% file: sec/5_conclusion.tex
\section{Conclusion}
\label{sec:conclusion}

We presented \emph{WorldSmith}, a single agent framework that creates and modifies 3D and 4D content by writing Blender programs, calling foundation models as tools, and checking renders of its own result. Without task-specific training, the same loop covers multi-turn asset editing, surface emission, rig-based animation, and 3D and
4D scene generation; per-task harnesses keep each change local and verifiable, from an issue ledger with preservation
checks for editing to native-rig keyframes for animation and layout-first construction for scenes. Across established benchmarks and our CompoScene-3D and CompoScene-4D benchmarks, the framework achieves the best reported results on key metrics. Comparisons with direct agents using the same model suggest that the framework contributes to the gains in asset editing and 3D scene generation.

%% file: sec/X_suppl.tex
\clearpage
\appendix
\setcounter{figure}{0}
\setcounter{table}{0}
\renewcommand{\thefigure}{S\arabic{figure}}
\renewcommand{\thetable}{S\arabic{table}}
\renewcommand{\theHfigure}{supp.\arabic{figure}}
\renewcommand{\theHtable}{supp.\arabic{table}}
\setlength{\textfloatsep}{8pt plus 2pt minus 2pt}
\setlength{\dbltextfloatsep}{8pt plus 2pt minus 2pt}
\setlength{\floatsep}{6pt plus 2pt minus 2pt}
\setlength{\dblfloatsep}{6pt plus 2pt minus 2pt}
\setlength{\abovecaptionskip}{4pt}
\setlength{\belowcaptionskip}{0pt}
\renewcommand{\dbltopfraction}{0.85}
\renewcommand{\topfraction}{0.85}
\renewcommand{\dblfloatpagefraction}{0.9}
\renewcommand{\floatpagefraction}{0.9}
\makeatletter
\setlength{\@fptop}{0pt}
\setlength{\@fpsep}{10pt plus 2pt}
\setlength{\@fpbot}{0pt plus 1fil}
\setlength{\@dblfptop}{0pt}
\setlength{\@dblfpsep}{10pt plus 2pt}
\setlength{\@dblfpbot}{0pt plus 1fil}
\makeatother

\twocolumn[{%
{\centering\Large
\textbf{\thetitle}\par
\vspace{0.5em}
Supplementary Material\par}
\vspace{0.5em}
\noindent
\begin{minipage}{\textwidth}
\centering
\captionsetup{type=table,position=top}
\caption{Quantitative results on Eval3DEdit with five-category macro averaging. Best values are bold; second-best underlined.}
\label{tab:eval3dedit}
\setlength{\tabcolsep}{4pt}
\small
\begin{tabular*}{\linewidth}{@{\extracolsep{\fill}}lcccc}
\toprule
Method & CLIP$_{\rm img}$ $\uparrow$ & CLIP$_{\rm txt}$ $\uparrow$ & Uni3D$_{\rm pc}$ $\uparrow$ & Uni3D$_{\rm txt}$ $\uparrow$ \\
\midrule
3DEditFormer~\cite{threeditformer} & 0.7185 & 0.4820 & 0.7918 & 0.2522 \\
PartFlow~\cite{partflow} & 0.7223 & 0.4815 & \underline{0.8448} & 0.2538 \\
Alchemy3D~\cite{alchemy3d} & \underline{0.7439} & \underline{0.4991} & 0.7823 & 0.2581 \\
Codex agent (GPT-6 Astra) & 0.7332 & 0.4934 & 0.8358 & \underline{0.2600} \\
\textbf{\ours} & \textbf{0.7469} & \textbf{0.5008} & \textbf{0.8548} & \textbf{0.2636} \\
\bottomrule
\end{tabular*}
\end{minipage}\par
\vspace{0.75em}
\noindent
\begin{minipage}{\textwidth}
\centering
\captionsetup{type=table,position=top}
\caption{Quantitative results on Steer3D Edit3D-Bench. PartFlow and 3DEditFormer support texture edits only; ``--'' denotes unavailable results. Best values are bold and second-best underlined within each block, ranked before rounding.}
\label{tab:steer3d_repository_code}
\setlength{\tabcolsep}{4pt}
\small
\begin{tabular*}{\linewidth}{@{\extracolsep{\fill}}lccccccccc}
\toprule
Method & \multicolumn{3}{c}{Removal} & \multicolumn{3}{c}{Addition} & \multicolumn{3}{c}{Texture} \\
\cmidrule(lr){2-4}\cmidrule(lr){5-7}\cmidrule(lr){8-10}
 & LPIPS $\downarrow$ & CD $\downarrow$ & F1 $\uparrow$ & LPIPS $\downarrow$ & CD $\downarrow$ & F1 $\uparrow$ & LPIPS $\downarrow$ & CD $\downarrow$ & F1 $\uparrow$ \\
\midrule
3DEditFormer~\cite{threeditformer} & -- & -- & -- & -- & -- & -- & 0.1475 & 0.0165 & 0.6734 \\
PartFlow~\cite{partflow} & -- & -- & -- & -- & -- & -- & \underline{0.1401} & \underline{0.0094} & 0.7048 \\
Nano3D~\cite{nano3d} & \underline{0.1204} & \underline{0.0128} & \underline{0.6539} & \underline{0.1771} & \underline{0.0096} & \textbf{0.6069} & 0.1433 & 0.0100 & \underline{0.7118} \\
Alchemy3D~\cite{alchemy3d} & 0.1467 & 0.0302 & 0.5564 & 0.1871 & 0.0205 & 0.5091 & 0.1786 & 0.0203 & 0.5765 \\
Codex agent (GPT-6 Astra) & 0.2244 & 0.0393 & 0.6215 & 0.2190 & 0.0213 & 0.5405 & 0.2308 & 0.0417 & 0.6255 \\
\textbf{\ours} & \textbf{0.0880} & \textbf{0.0040} & \textbf{0.7552} & \textbf{0.1540} & \textbf{0.0096} & \underline{0.5947} & \textbf{0.1217} & \textbf{0.0093} & \textbf{0.7409} \\
\bottomrule
\end{tabular*}
\end{minipage}\par
\vspace{0.75em}
\noindent
\begin{minipage}{\textwidth}
\centering
\captionsetup{type=table,position=top}
\caption{Quantitative results on VoxHammer Edit3D-Bench (300 editing tasks). CD and masked image metrics compare unedited regions against the source. Best values are bold; second-best underlined.}
\label{tab:voxhammer}
\setlength{\tabcolsep}{4pt}
\small
\begin{tabular*}{\linewidth}{@{\extracolsep{\fill}}lcccccc}
\toprule
Method & CD $\downarrow$ & PSNR$_M$ $\uparrow$ & SSIM$_M$ $\uparrow$ & LPIPS$_M$ $\downarrow$ & FID $\downarrow$ & CLIP-T $\uparrow$ \\
\midrule
VoxHammer & 0.0120 & 41.6800 & \underline{0.9940} & 0.0270 & \underline{23.0500} & \underline{0.2870} \\
Nano3D~\cite{nano3d} & \underline{0.0101} & 31.4200 & 0.9750 & 0.0350 & 24.1200 & 0.2800 \\
3DEditFormer~\cite{threeditformer} & 0.0213 & 27.4400 & 0.9480 & 0.0610 & 28.0300 & 0.2780 \\
PartFlow~\cite{partflow} & 0.0153 & 30.1300 & 0.9670 & 0.0440 & 25.3300 & 0.2770 \\
Alchemy3D~\cite{alchemy3d} & 0.0226 & 31.9113 & 0.9675 & 0.0439 & 26.1726 & 0.2868 \\
Codex agent (GPT-6 Astra) & 0.1565 & \textbf{55.7033} & 0.9932 & \underline{0.0226} & 26.4200 & 0.2810 \\
\textbf{\ours} & \textbf{0.0073} & \underline{54.7667} & \textbf{0.9987} & \textbf{0.0166} & \textbf{22.9182} & \textbf{0.2892} \\
\bottomrule
\end{tabular*}
\end{minipage}\par
\vspace{0.75em}
}]

\section{Additional 3D Editing Benchmarks}
We report additional evaluations on three single-turn 3D editing benchmarks: Eval3DEdit (\cref{tab:eval3dedit}), Steer3D Edit3D-Bench (\cref{tab:steer3d_repository_code}), and VoxHammer Edit3D-Bench (\cref{tab:voxhammer}).

On Eval3DEdit (\cref{tab:eval3dedit}), metrics are macro-averaged across five edit categories. \ours{} leads all compared methods on all four metrics, outperforming both specialized editors and the same-model Codex agent baseline. Scored coverage is 99/100 for 3DEditFormer and 98/100 for the Codex agent.

On Steer3D Edit3D-Bench (\cref{tab:steer3d_repository_code}), \ours{} achieves the best score on 8 of 9 metrics across removal, addition, and texture edits, with the largest margins on removal (CD $0.0040$ vs.\ $0.0128$ and F1 $0.7552$ vs.\ $0.6539$ for Nano3D). PartFlow and 3DEditFormer report texture-only results, and Alchemy3D completes 248 of 250 tasks.

On VoxHammer Edit3D-Bench (\cref{tab:voxhammer}), published baseline numbers are transcribed from their respective papers while local runs are evaluated under the benchmark protocol (with FID computed over the corresponding prediction set). \ours{} achieves the best CD ($0.0073$), SSIM$_M$ ($0.9987$), LPIPS$_M$ ($0.0166$), FID ($22.9182$), and CLIP-T ($0.2892$). Because Blender-based local mesh edits preserve unedited faces directly without re-decoding the entire mesh, both agentic methods attain substantially higher masked PSNR$_M$ on unedited regions than neural regeneration baselines; however, the unconstrained Codex agent suffers severe geometric distortion (CD $0.1565$), whereas \ours{} maintains both high preservation and geometric accuracy.
\begin{table*}[t]
\centering
\caption{Self-rollout results on EditHero by edit type (457 chains, 2{,}755 edits). Best scores are bold, second-best underlined.}
\label{tab:edithero_by_type}
\setlength{\tabcolsep}{2pt}
\small
\begin{tabular*}{\linewidth}{@{\extracolsep{\fill}}lcccccccccccc}
\toprule
Method
 & \multicolumn{3}{c}{Removal (510)}
 & \multicolumn{3}{c}{Addition (968)}
 & \multicolumn{3}{c}{Replacement (629)}
 & \multicolumn{3}{c}{Retexture (648)} \\
\cmidrule(lr){2-4}\cmidrule(lr){5-7}\cmidrule(lr){8-10}\cmidrule(lr){11-13}
 & LPIPS $\downarrow$ & CD $\downarrow$ & F $\uparrow$
 & LPIPS $\downarrow$ & CD $\downarrow$ & F $\uparrow$
 & LPIPS $\downarrow$ & CD $\downarrow$ & F $\uparrow$
 & LPIPS $\downarrow$ & CD $\downarrow$ & F $\uparrow$ \\
\midrule
3DEditFormer~\cite{threeditformer}
 & 0.209 & 3.01 & 0.546
 & 0.263 & 4.37 & 0.421
 & 0.191 & 2.28 & 0.558
 & 0.173 & 1.81 & 0.625 \\
PartFlow~\cite{partflow}
 & 0.185 & 2.33 & 0.664
 & 0.233 & 3.39 & 0.578
 & 0.178 & 1.74 & 0.640
 & 0.141 & 1.10 & 0.744 \\
Nano3D~\cite{nano3d}
 & 0.130 & 1.63 & 0.745
 & 0.225 & 3.58 & 0.604
 & 0.149 & 1.43 & 0.686
 & 0.143 & 0.91 & 0.801 \\
Alchemy3D~\cite{alchemy3d}
 & 0.204 & 4.48 & 0.443
 & 0.313 & 7.00 & 0.272
 & 0.204 & 3.28 & 0.479
 & 0.163 & 1.71 & 0.675 \\
Codex agent (GPT-6 Astra)
 & \underline{0.027} & \underline{0.56} & \underline{0.905}
 & \underline{0.083} & \textbf{1.08} & \underline{0.751}
 & \underline{0.066} & \underline{0.83} & \underline{0.822}
 & \underline{0.055} & \underline{0.57} & \underline{0.905} \\
\textbf{\ours}
 & \textbf{0.024} & \textbf{0.56} & \textbf{0.909}
 & \textbf{0.080} & \underline{1.14} & \textbf{0.759}
 & \textbf{0.057} & \textbf{0.78} & \textbf{0.837}
 & \textbf{0.043} & \textbf{0.56} & \textbf{0.908} \\
\bottomrule
\end{tabular*}
\end{table*}
\section{Component Ablations}

\subsection{3D Editing: Issue Ledger}
\begin{table*}[t]
\centering
\caption{Issue-ledger ablation on an EditHero subset (14 randomly selected turns). $E$ and $U$ denote the edited and unedited regions, respectively, and LPIPS$_E^{\mathrm{held}}$ measures edited-region LPIPS on held-out views. Deltas are computed before rounding.}
\label{tab:edithero-issue-ledger-add}
\small
\setlength{\tabcolsep}{4pt}
\begin{tabular*}{\textwidth}{@{\extracolsep{\fill}}lcccc}
\toprule
Variant & IF $\uparrow$ & LPIPS$_E$ $\downarrow$ & LPIPS$_E^{\mathrm{held}}$ $\downarrow$ & LPIPS$_U$ $\downarrow$ \\
\midrule
Full & \textbf{0.4842} & \textbf{0.2098} & \textbf{0.2361} & \textbf{0.0406} \\
Without issue ledger & 0.4683 & 0.2413 & 0.2672 & 0.0440 \\
\midrule
$\Delta$ (Without $-$ Full) & $-0.0159$ & $+0.0315$ & $+0.0311$ & $+0.0034$ \\
\bottomrule
\end{tabular*}
\end{table*}

\Cref{tab:edithero-issue-ledger-add} ablates the issue ledger on 14 randomly sampled EditHero turns. The ablated variant removes issue-ledger updates, current-edit focus, and ledger-derived capability hints while retaining fresh candidate evidence, editing tools, static delivery, and source-preservation checks. Removing the issue ledger degrades every metric: instruction following drops by $0.0159$ ($0.4842 \to 0.4683$), edited-region perceptual error increases on both the conditioning view ($\text{LPIPS}_E$: $+0.0315$) and held-out views ($\text{LPIPS}_E^{\mathrm{held}}$: $+0.0311$), and unedited-region drift also worsens ($\text{LPIPS}_U$: $+0.0034$). This confirms that tracking prioritized, evidence-bound issues helps the agent converge to more accurate local edits without introducing unintended changes.

\subsection{3D Scene Generation}

\begin{table*}[tp]
\centering
\caption{Mesh selection and silhouette fitting ablation on all 120 \benchthree{} scenes. All variants produce 120 scenes. S-IoU and O-IoU denote scene and object IoU, with A and O denoting AABB and OBB. PL is $100\times$ normalized RGB MSE.}
\label{tab:scene-ablation}
\small
\begin{tabular*}{\textwidth}{@{\extracolsep{\fill}}lrrrrrr@{}}
\toprule
Variant & S-IoU$_A$ $\uparrow$ & S-IoU$_O$ $\uparrow$ & O-IoU$_A$ $\uparrow$ & O-IoU$_O$ $\uparrow$ & PL $\downarrow$ & N-CLIP $\downarrow$ \\
\midrule
All low-poly (layout only) & 0.8243 & 0.5781 & 0.4426 & 0.3542 & 1.4057 & 0.9423 \\
All high-poly & 0.8443 & 0.5966 & 0.4681 & 0.3795 & 1.2494 & 0.8249 \\
Full (review-selected) & 0.8357 & 0.5889 & 0.4587 & 0.3720 & 1.2996 & 0.8345 \\
Full without silhouette fitting & 0.8030 & 0.5789 & 0.4078 & 0.3334 & 1.5081 & 0.9874 \\
\bottomrule
\end{tabular*}
\end{table*}

\paragraph{Layout, silhouette fitting, and mesh review.}
\Cref{tab:scene-ablation} evaluates the layout, mesh-selection, and silhouette-fitting stages across all 120 \benchthree{} scenes (confidence intervals use 10{,}000 paired scene bootstrap draws). Even the low-poly layout alone (O-IoU$_A$ $0.4426$) substantially outperforms the direct GPT-6 Astra baseline in the main paper ($0.399$), confirming that layout-first primitive construction drives instance placement accuracy. Adding generated meshes improves both geometry and appearance (PL $1.4057 \to 1.2996$, N-CLIP $0.9423 \to 0.8345$).

To isolate silhouette fitting, we fix the generated meshes and review decisions, place each accepted mesh at the proxy's bounding-box center using the initializer's uniform scale, and skip yaw search and silhouette minimization. Silhouette fitting improves Object AABB IoU by $+0.0509$ (95\% CI $[0.0357,0.0679]$) and Object OBB IoU by $+0.0386$ ($[0.0268,0.0512]$), while reducing PL from $1.5081$ to $1.2996$ and N-CLIP from $0.9874$ to $0.8345$.

\begin{figure*}[tp]
\centering
\includegraphics[width=0.9\textwidth]{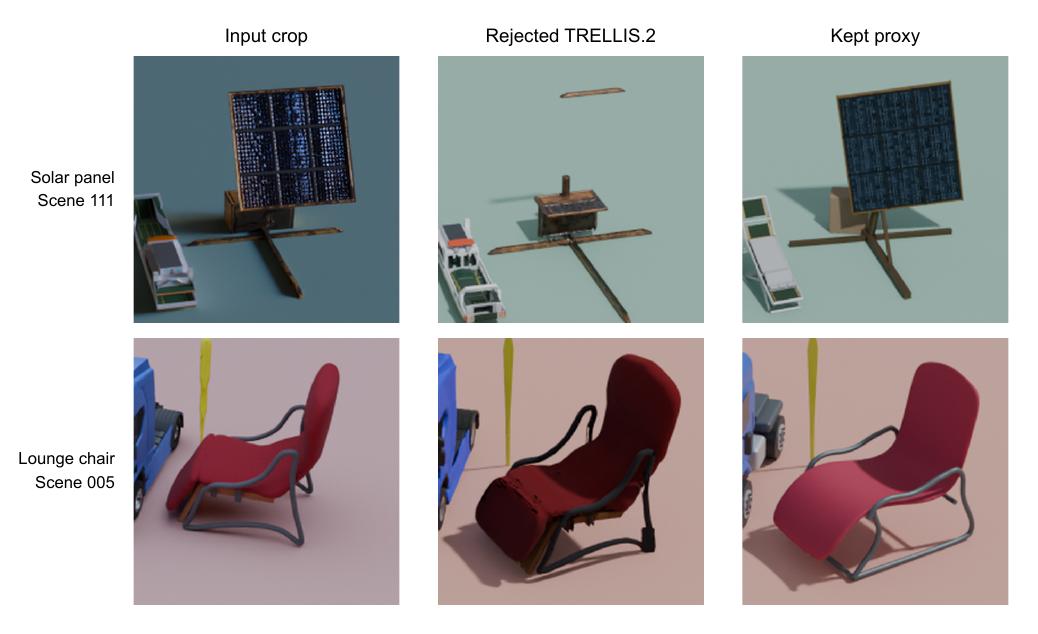}
\caption{Two examples where the mesh reviewer retained the low-poly proxy. The high-poly solar-panel candidate lacks the panel surface and contains floating fragments; the chair candidate exhibits a distorted seat edge and exposed supports.}
\label{fig:mesh-review-fallbacks}
\end{figure*}

Comparing Full (review-selected) with All high-poly, keeping every generated mesh yields slightly higher point estimates across all six metrics, though every paired 95\% confidence interval between the two variants spans zero. Overall, the reviewer retained the low-poly proxy for 240 of 442 assessed candidates (54.3\%); four additional candidates exceeded the per-scene review budget and defaulted to the proxy, and 22 objects had no fitted candidate (266 of 468 proxies retained in total). We do not default to high-poly meshes for all objects for two reasons. First, high-poly mesh generation is not uniformly reliable: some objects fail to yield a fitted candidate, while others exhibit localized structural defects---such as missing surfaces, floating fragments, or distorted parts (\cref{fig:mesh-review-fallbacks})---that coarse box IoU and global pixel MSE do not penalize. Second, using high-poly meshes across every object increases computational and asset cost: dense meshes substantially increase polygon counts, memory footprint, and rendering or downstream editing overhead, and regenerating failed or defective candidates to enforce all-high-poly scenes would further increase generation cost, whereas lightweight low-poly proxies already suffice for simpler geometries. The reviewer therefore acts as a conservative visual and cost-aware safeguard that adopts high-poly assets only when they reliably improve upon the proxy.

\begin{table*}[tp]
\centering
\caption{Occlusion completion versus direct RGB cropping on 20 randomly sampled \benchthree{} scenes. The occluded subset comprises 14 matched objects across 11 scenes. Differences are completion minus direct crop, with 95\% CIs from 10{,}000 paired scene bootstrap draws.}
\label{tab:scene-completion-ablation}
\small
\begin{tabular*}{\textwidth}{@{\extracolsep{\fill}}llrrr@{}}
\toprule
Subset & Metric & Direct crop & Completion & $\Delta$ [95\% CI] \\
\midrule
All 20 scenes & Object AABB IoU $\uparrow$ & 0.2854 & 0.4140 & $+0.1287$ $[0.0656,0.1973]$ \\
 & Object OBB IoU $\uparrow$ & 0.2390 & 0.3367 & $+0.0978$ $[0.0441,0.1559]$ \\
 & PL $\downarrow$ & 2.2265 & 1.8324 & $-0.3941$ $[-0.7857,-0.0432]$ \\
 & N-CLIP $\downarrow$ & 1.1086 & 0.8973 & $-0.2113$ $[-0.3649,-0.0557]$ \\
\midrule
Occluded objects & AABB IoU $\uparrow$ & 0.1819 & 0.3538 & $+0.1719$ $[0.0557,0.3066]$ \\
 & OBB IoU $\uparrow$ & 0.1715 & 0.2722 & $+0.1007$ $[0.0134,0.2005]$ \\
\bottomrule
\end{tabular*}
\end{table*}

\paragraph{Occlusion completion.}
\Cref{tab:scene-completion-ablation} ablates proxy-guided reference completion on 20 randomly sampled \benchthree{} scenes (sampled without replacement). Both settings share identical layouts, crop boxes, TRELLIS.2 hyperparameters and seeds, and silhouette fitting, with mesh review disabled so all fitted meshes are evaluated directly (two objects without valid crops retained their proxies in both settings). The direct baseline passes the raw $512^2$ RGB crop to TRELLIS.2 using its default foreground preprocessing, whereas the completion setting uses proxy-filled crops completed with \texttt{gpt-image-2}. The BiRefNet mask IoU gate ($\geq 0.55$ against the pre-completion mask) accepted 67 of 70 completions, with the remaining three falling back to the proxy-filled crop.

Across all 20 scenes, proxy-guided completion yields statistically significant gains on every metric, raising Object AABB IoU by $+0.1287$ and Object OBB IoU by $+0.0978$ while lowering PL ($-0.3941$) and N-CLIP ($-0.2113$). Defining an occluded object as one whose visible area is at most 80\% of its isolated proxy silhouette under the input camera (fixing ground-truth matches from the layout evaluation and averaging within each scene), the AABB IoU gain grows to $+0.1719$ ($0.1819 \to 0.3538$).

\subsection{4D Scene Generation}

\begin{table*}[tp]
\centering
\caption{Static versus motion-enabled reconstruction on \benchfour{}. Missing predictions (3 of 120 scenes) are scored as zero for F-score, while CD and trajectory error are averaged over the scored pairs. Differences are motion-enabled minus static, with 95\% CIs from 10{,}000 paired bootstrap draws clustered by source character.}
\label{tab:4d-ablation}
\small
\begin{tabular*}{\textwidth}{@{\extracolsep{\fill}}llrrrr@{}}
\toprule
Subset ($N$) & Metric & Pairs & Static & Motion-enabled & $\Delta$ [95\% CI] \\
\midrule
All (120) & World F@5\% $\uparrow$ & 117 & 0.4411 & 0.4502 & $+0.0092$ $[-0.0087,0.0266]$ \\
 & World CD $\downarrow$ & 117 & 0.2331 & 0.2272 & $-0.0059$ $[-0.0273,0.0160]$ \\
 & Local CD $\downarrow$ & 117 & 0.2168 & 0.2015 & $-0.0153$ $[-0.0392,0.0034]$ \\
 & Center trajectory error $\downarrow$ & 117 & 0.0776 & 0.0734 & $-0.0042$ $[-0.0142,0.0040]$ \\
 & Peak-frame World F@5\% $\uparrow$ & 117 & 0.3777 & 0.4101 & $+0.0325$ $[0.0086,0.0568]$ \\
 & Peak-frame World CD $\downarrow$ & 117 & 0.2774 & 0.2512 & $-0.0261$ $[-0.0649,0.0024]$ \\
\midrule
Small motion (60) & World F@5\% $\uparrow$ & 58 & 0.4675 & 0.4785 & $+0.0110$ $[-0.0113,0.0312]$ \\
 & World CD $\downarrow$ & 58 & 0.1953 & 0.2039 & $+0.0085$ $[-0.0124,0.0438]$ \\
 & Peak-frame World F@5\% $\uparrow$ & 58 & 0.4377 & 0.4740 & $+0.0363$ $[0.0027,0.0735]$ \\
 & Peak-frame World CD $\downarrow$ & 58 & 0.2030 & 0.2037 & $+0.0007$ $[-0.0216,0.0342]$ \\
\midrule
Large motion (60) & World F@5\% $\uparrow$ & 59 & 0.4146 & 0.4220 & $+0.0073$ $[-0.0223,0.0345]$ \\
 & World CD $\downarrow$ & 59 & 0.2702 & 0.2501 & $-0.0201$ $[-0.0537,0.0042]$ \\
 & Peak-frame World F@5\% $\uparrow$ & 59 & 0.3176 & 0.3463 & $+0.0287$ $[-0.0058,0.0625]$ \\
 & Peak-frame World CD $\downarrow$ & 59 & 0.3505 & 0.2980 & $-0.0525$ $[-0.1219,-0.0086]$ \\
\midrule
Animated (106) & World F@5\% $\uparrow$ & 106 & 0.4531 & 0.4635 & $+0.0104$ $[-0.0104,0.0291]$ \\
 & Peak-frame World F@5\% $\uparrow$ & 106 & 0.3897 & 0.4265 & $+0.0368$ $[0.0089,0.0654]$ \\
\bottomrule
\end{tabular*}
\end{table*}

\paragraph{Pipeline reliability and motion component.}
We break down per-stage reliability across the 120 \benchfour{} scenes (117 produced static reconstructions; 3 failed). The temporal-change score in Eq.~(3) correctly identified the ground-truth moving character in 111 of 117 reconstructed scenes (94.9\%, or 92.5\% of all 120 scenes), evaluated by matching the selected proxy to the ground-truth instance with the highest positive first-frame mask IoU. SAM~2 produced complete tracking masks for 112 of 117 scenes, and SkinTokens plus Puppeteer succeeded in rigging 106 of those 112 scenes, which proceeded to animation and assembly. Five tracking failures and six rigging failures fell back cleanly to the static reconstruction. Under the evaluator's first-frame Hungarian instance matching, the animated character was paired with the ground-truth character in 95 of the 106 animated scenes.

\Cref{tab:4d-ablation} compares the full motion-enabled output against keeping the static first-frame reconstruction frozen across all frames. To separate dynamic deformation from near-rest frames, we stratify the 120 scenes into equal small-motion and large-motion halves (60 scenes each) ranked by the maximum frame-wise RMS displacement of ground-truth vertices relative to frame one (normalized by the largest side of the first-frame bounding box), and report metrics both averaged over all frames and at each scene's ground-truth peak-motion frame. Center trajectory error averages the normalized distance between predicted and ground-truth AABB center displacements from frame one. Because all-frame averages are diluted by frames near the rest pose and by small-amplitude clips, all-frame World F@5\% differences span zero. At the peak-motion frame, however, enabling animation yields statistically significant improvements in World F@5\% across both the full split ($+0.0325$, 95\% CI $[0.0086,0.0568]$) and the 106 animated scenes ($+0.0368$, $[0.0089,0.0654]$), and substantially reduces peak-frame World CD on large-motion scenes ($-0.0525$, $[-0.1219,-0.0086]$).

\begin{table*}[tp]
\centering
\caption{Ablation of first-frame Umeyama alignment in Eq.~(4) on \benchfour{}. Differences are with minus without alignment, using the same evaluation protocol and 95\% bootstrap CIs as \cref{tab:4d-ablation}.}
\label{tab:4d-anchor-ablation}
\small
\begin{tabular*}{\textwidth}{@{\extracolsep{\fill}}llrrr@{}}
\toprule
Subset & Metric & Without & With & $\Delta$ [95\% CI] \\
\midrule
Full split (120, 117 scored) & World F@5\% $\uparrow$ & 0.3821 & 0.4502 & $+0.0682$ $[0.0416,0.0966]$ \\
 & World CD $\downarrow$ & 0.2717 & 0.2272 & $-0.0446$ $[-0.0784,-0.0184]$ \\
Assembled actor (106) & World F@5\% $\uparrow$ & 0.3785 & 0.4503 & $+0.0718$ $[0.0384,0.1054]$ \\
 & World CD $\downarrow$ & 0.3241 & 0.2684 & $-0.0557$ $[-0.0993,-0.0195]$ \\
\bottomrule
\end{tabular*}
\end{table*}

\paragraph{First-frame Umeyama alignment.}
\Cref{tab:4d-anchor-ablation} ablates the Umeyama similarity transform in Eq.~(4) by placing the same animated mesh back into the scene using $C^{-1}$ alone while holding scene alignment and instance assignments fixed. We also report direct actor-level evaluation across all 106 assembled animated characters (including the 11 scenes where the benchmark evaluator matched a different predicted instance on frame one). Anchoring the first animated frame with Eq.~(4) produces large, statistically significant gains on both splits, increasing World F@5\% by $+0.0682$ ($+0.0718$ on assembled actors) and reducing World CD by $0.0446$ ($0.0557$ on assembled actors).

\paragraph{Camera parameters for 3D and 4D scenes.}
Across both 3D and 4D scene generation (\benchthree{} and \benchfour{}), all experiments use the ground-truth camera parameters (intrinsics and extrinsics) for both \ours{} and the compared baselines. Specifically, the same ground-truth camera is provided to all baselines whose pipelines accept camera parameters (including direct coding agents and camera-conditioned reconstruction methods), and is used identically across all methods during rendering and evaluation. Using shared ground-truth camera parameters ensures a fair comparison of scene layout, geometry, appearance, and motion under identical camera conditions.

\subsection{Surface Emission}

\begin{table*}[tp]
\centering
\caption{Surface emission ablations on LightGenBench (official evaluator, 50{,}000 surface samples per shape). The top block evaluates fusion, view coverage, and color reuse on the full 200-shape test split; the bottom block evaluates iterative mask refinement on 30 randomly sampled test shapes.}
\label{tab:emission-component-ablation}
\small
\begin{tabular*}{\textwidth}{@{\extracolsep{\fill}}lrrrr@{}}
\toprule
Variant & $N$ & IoU $\uparrow$ & MAE $\downarrow$ & PSNR (dB) $\uparrow$ \\
\midrule
Weighted UV fusion, all available views, base color & 200 & 0.3378 & 0.0266 & 27.23 \\
Equal-weight majority, all available views, base color & 200 & 0.3321 & 0.0264 & 27.36 \\
Union, all available views, base color & 200 & 0.3310 & 0.0270 & 27.13 \\
Weighted UV fusion, four base views, base color & 200 & 0.3069 & 0.0267 & 27.11 \\
Weighted UV fusion, all available views, white & 200 & 0.3378 & 0.0634 & 20.40 \\
\midrule
First segmentation per view (unrefined) & 30 & 0.2760 & 0.0433 & 24.59 \\
Final refined masks & 30 & 0.3790 & 0.0196 & 26.83 \\
\bottomrule
\end{tabular*}
\end{table*}

\Cref{tab:emission-component-ablation} ablates the components of our surface emission harness (all 95\% CIs use 10{,}000 paired shape bootstrap draws). In the full-split comparison ($N=200$), both majority voting (excluding ties) and union operate only over views that observe each texel. Weighted UV fusion in Eq.~(2) improves IoU by $+0.0069$ over union (95\% CI $[0.0026,0.0120]$) and $+0.0058$ over equal-weight majority ($[-0.0023,0.0159]$). Re-fusing the saved per-view masks offline yields IoU $0.3378$ (compared to $0.3388$, reported as $0.339$ in the main paper) due to two shapes where online timeout fallback rules differed. Adding extra cameras beyond the four base views improves IoU by $+0.0309$ ($[0.0137,0.0526]$) across the full split, and by $+0.1586$ ($[0.0765,0.2562]$) on the 39 shapes that required additional views to cover occluded surfaces. At fixed mask support, reusing base color $a(t)$ rather than emitting constant white reduces MAE by $0.0368$ ($[0.0251,0.0500]$) and increases PSNR by $+6.83$\,dB ($[5.89,7.79]$).

To isolate iterative visual refinement, we evaluate 30 randomly sampled test shapes using identical camera views, projection, weighted UV fusion, and base-color export. The unrefined baseline takes the initial segmentation mask returned in each view (first SAM~3 detection union, highest-scoring first-call SAM~2 mask, or initial color mask) before zoomed boundary inspection and prompt/exclusion refinement. Iterative refinement improves IoU by $+0.1029$ ($[0.0364,0.1860]$), reduces MAE by $0.0237$ ($[0.0020,0.0618]$), and raises PSNR by $+2.24$\,dB ($[0.68,4.21]$).

\subsection{Rig-based Animation}

\begin{table*}[tp]
\centering
\caption{Candidate iteration and selection ablation on Motion-80.}
\label{tab:motion80-candidate-ablation}
\small
\begin{tabular*}{\textwidth}{@{\extracolsep{\fill}}lrrrr@{}}
\toprule
 & \multicolumn{2}{c}{79 subjects, GLB} & \multicolumn{2}{c}{All 80 subjects} \\
\cmidrule(lr){2-3}\cmidrule(lr){4-5}
Candidate or selection rule & CD $\downarrow$ & F$_{0.02}$ $\uparrow$ & CD $\downarrow$ & F$_{0.02}$ $\uparrow$ \\
\midrule
Candidate 1 (first complete draft) & 0.1499 & 0.3194 & 0.1520 & 0.3163 \\
Candidate 2 & 0.1260 & 0.3414 & 0.1275 & 0.3415 \\
Candidate 3 & 0.1239 & 0.3335 & 0.1248 & 0.3348 \\
Candidate 4 & 0.1249 & 0.3447 & 0.1265 & 0.3448 \\
Mean of candidates 2--4 & 0.1249 & 0.3398 & 0.1263 & 0.3403 \\
Uniform draw from the comparison pool (expected) & 0.1241 & 0.3487 & 0.1254 & 0.3491 \\
Last saved comparison candidate & 0.1157 & 0.3706 & 0.1170 & 0.3705 \\
Final agent delivery & 0.1163 & 0.3694 & 0.1179 & 0.3691 \\
\bottomrule
\end{tabular*}
\end{table*}

\Cref{tab:motion80-candidate-ablation} evaluates candidate progression and blind A/B review on Motion-80 using the official first-frame alignment and 50{,}000 surface samples per frame (95\% CIs from 10{,}000 paired subject bootstrap draws). Ordering saved candidates chronologically, the random baseline averages uniformly over each subject's 4--11 scored candidates. On the standard 79-subject GLB subset, iterative refinement reduces CD by $0.0336$ (95\% CI $[0.0190,0.0515]$) and increases F$_{0.02}$ by $+0.0500$  ($[0.0332,0.0697]$), and also beats uniform random selection from the candidate pool ($-0.0078$ CD, $[-0.0114,-0.0045]$; $+0.0207$ F$_{0.02}$, $[0.0127,0.0310]$). Because each subsequent revision is authored using feedback from prior blind reviews, the last saved candidate already accumulates the reviewer's guidance and achieves virtually identical geometry to the final delivered candidate ($0.1157$ vs.\ $0.1163$ CD).

We further evaluate the blind A/B reviewer directly across all 467 recorded pairwise comparisons from the 80 subjects (418 with an explicit preference; ties resolved by uniform draw, averaged within each subject before aggregation). Relative to random selection on the exact same candidate pairs, the blind reviewer improves CD from $0.1287$ to $0.1252$ ($\Delta = -0.0034$, 95\% CI $[-0.0065,0.0005]$) and significantly improves F$_{0.02}$ from $0.3457$ to $0.3519$ ($\Delta = +0.0061$, $[0.0022,0.0099]$).

\section{Computational Cost}

\begin{table*}[tp]
\centering
\caption{Mean computational cost per benchmark unit across all five tasks. Superscripts denote image-completion requests ($^{\rm I}$), TRELLIS.2 mesh-generation calls ($^{\rm T}$), summed reconstruction stage durations ($^{\rm S}$), and animation-stage duration through delivery ($^{\rm A}$). For 4D scenes, author turns and logged tools split layout and animation stages ($+$). Token totals include cached input tokens where reported.}
\label{tab:task-runtime-cost}
\small
\setlength{\tabcolsep}{3pt}
\begin{tabular*}{\textwidth}{@{\extracolsep{\fill}}llrrrrr@{}}
\toprule
Task & Units & Author turns & Model requests & Logged tools & Minutes & Tokens \\
\midrule
3D editing & 2{,}755 edits & 30.96 & 30.96 & 31.58 & 13.58 & 683{,}711 \\
Surface emission & 200 shapes & 10.96 & 10.97 & 18.99 & 6.41 & 301{,}000 \\
Rig-based animation & 80 subjects & 34.29 & 44.38 & 46.91 & 46.98 & 1{,}280{,}036 \\
3D scenes & 120 scenes & 4.62 & $12.78+3.72^{\rm I}$ & $3.72^{\rm T}$ & $21.35^{\rm S}$ & 68{,}021 \\
4D scenes & 106 assembled & $4.99+45.42$ & $73.94+3.28^{\rm I}$ & $66.38+3.32^{\rm T}$ & $87.86^{\rm A}$ & 1{,}838{,}746 \\
\bottomrule
\end{tabular*}
\end{table*}

\Cref{tab:task-runtime-cost} summarizes the average agent iterations, tool invocations, wall-clock duration, and token consumption per benchmark unit. Author turns count main-agent reasoning rounds (listed separately for layout and animation in 4D scenes), and model requests include reviewer calls. Logged tools count structured tool calls for editing and 4D animation, and saved tool outputs for emission and rig-based animation; host-side Blender utility calls are not separately totaled for 3D scenes, where token counts reflect language-model usage.

Editing time measures the delivered attempt per turn across all 457 chains, while emission and rig-based animation measure end-to-end execution through delivery. For 4D scenes, the static reconstruction stages averaged 29.32 minutes across 117 scenes, and across the 83 scenes with complete end-to-end timing records, static reconstruction, tracking, rigging, animation, and assembly together averaged 120.67 minutes (excluding inter-stage queue wait). We note two practical factors behind these figures: (1) multi-turn token totals accumulate cached prompt prefixes across revisions rather than newly generated tokens alone, and (2) wall-clock time is dominated by high-reasoning API latency, iterative Blender rendering, and sequential candidate verification. As such, \ours{} targets offline, high-precision authoring of editable 3D/4D assets and can be accelerated by parallelizing candidate generation or substituting faster reasoning models.

\section{Qualitative 3D Editing}

\begin{figure*}[t]
\centering
\includegraphics[width=\textwidth]{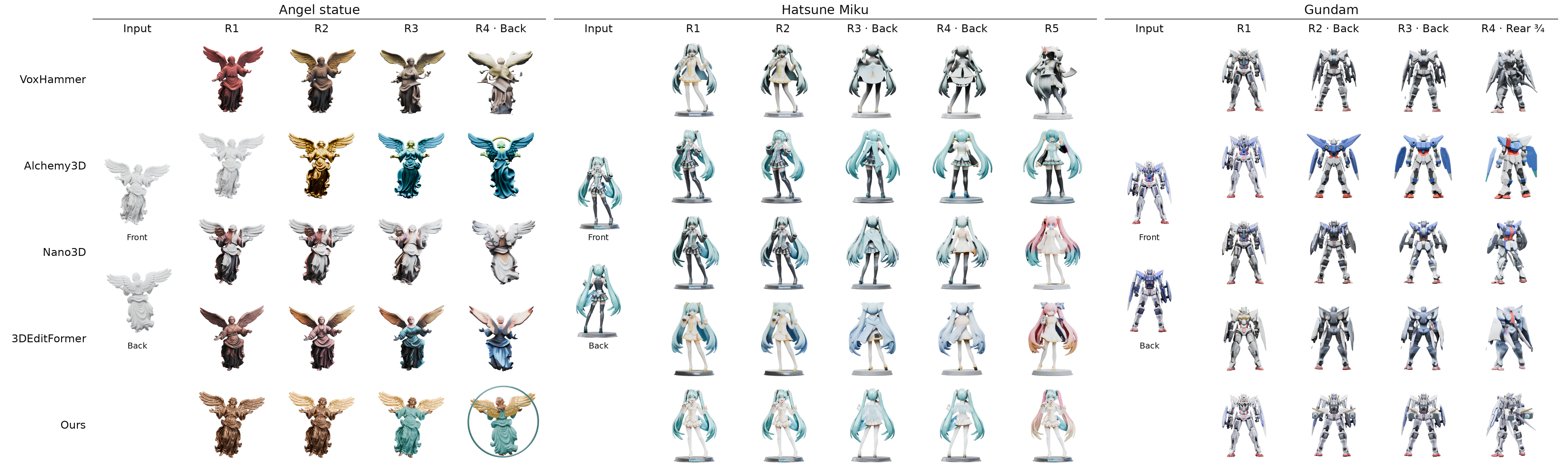}
\caption{Qualitative multi-turn 3D editing on three high-poly assets (angel statue, Hatsune Miku, Gundam): the input (front and back) and the results of VoxHammer, Alchemy3D, Nano3D, 3DEditFormer, and \ours{} after each edit request (R1, R2, \ldots; ``Back'' and ``Rear 3/4'' indicate the view shown).}
\label{fig:qualitative}
\end{figure*}

\Cref{fig:qualitative} shows multi-turn 3D editing results on three high-poly assets (an angel statue, Hatsune Miku, and a Gundam), comparing \ours{} against four reproduced baselines across both front and back views. While feedforward and latent-space baselines accumulate geometric distortion, surface artifacts, and texture drift after the initial turns, \ours{} applies each requested edit locally while preserving the unedited geometry and textures across all turns.

\section{Motion-80: Short and Long Sequences}
\Cref{tab:motion80_full} disaggregates the Motion-80 results from the main paper (\cref{tab:motion80}) into its short (64 subjects) and long (16 subjects) splits. \ours{} leads all compared methods on every metric across both splits, cutting Chamfer distance by nearly half on short sequences ($0.1145$ vs.\ $0.2090$ for MorphGS) and maintaining strong temporal coherence on long sequences (FVD $921.60$ vs.\ $1361.76$ for CAST; DreamSim $0.0845$ vs.\ $0.1381$ for MorphGS).

\begin{table*}[t]
\centering
\caption{Quantitative comparison on Motion-80, per split. All methods receive the rest mesh, original skeleton and skinning weights, and the video. SR is computed over the full split (64 short, 16 long); quality metrics are averaged over the subjects that all methods complete (61 short, 15 long), with appearance over three held-out views. $^{\dagger}$Uses a 512-joint input limit. Best in bold, second best underlined, within each split.}
\label{tab:motion80_full}
\small
\setlength{\tabcolsep}{4pt}
\begin{tabular*}{\textwidth}{@{\extracolsep{\fill}}lrrrrrrr}
\toprule
Method & SR (\%) $\uparrow$ & CD $\downarrow$ & F$_{0.02}$ $\uparrow$ & LPIPS $\downarrow$ & CLIP $\uparrow$ & FVD $\downarrow$ & DreamSim $\downarrow$ \\
\midrule
\multicolumn{8}{l}{\textit{Short sequences}} \\
CAST$^{\dagger}$~\cite{castmocap} & 100.00 & 0.2839 & 0.1797 & 0.1860 & 0.8562 & 1960.53 & 0.2146 \\
TopoCap~\cite{topocap} & 95.31 & 1.4712 & 0.1110 & 0.2032 & 0.8060 & 2344.98 & 0.2719 \\
MorphGS~\cite{morphgs} & 100.00 & \underline{0.2090} & \underline{0.2274} & \underline{0.1498} & \underline{0.8807} & \underline{1568.94} & \underline{0.1744} \\
MoCapAnything V2$^{\dagger}$~\cite{mocapv2} & 100.00 & 0.7525 & 0.1007 & 0.2145 & 0.7498 & 2541.84 & 0.3607 \\
\cmidrule(lr){1-8}
\textbf{\ours} & 100.00 & \textbf{0.1145} & \textbf{0.3933} & \textbf{0.1212} & \textbf{0.9460} & \textbf{893.16} & \textbf{0.0892} \\
\midrule
\multicolumn{8}{l}{\textit{Long sequences}} \\
CAST$^{\dagger}$ & 100.00 & \underline{0.2027} & 0.1773 & 0.1635 & \underline{0.9011} & \underline{1361.76} & 0.1448 \\
TopoCap & 93.75 & 0.6295 & 0.0788 & 0.2188 & 0.8442 & 2304.68 & 0.2197 \\
MorphGS & 100.00 & 0.2041 & \underline{0.1785} & \underline{0.1498} & 0.8984 & 1538.77 & \underline{0.1381} \\
MoCapAnything V2$^{\dagger}$ & 100.00 & 0.5420 & 0.0806 & 0.2124 & 0.7656 & 2483.28 & 0.3385 \\
\cmidrule(lr){1-8}
\textbf{\ours} & 100.00 & \textbf{0.1318} & \textbf{0.2683} & \textbf{0.1222} & \textbf{0.9416} & \textbf{921.60} & \textbf{0.0845} \\
\bottomrule
\end{tabular*}
\end{table*}

\section{Qualitative Scene Generation}
\begin{figure*}[t]
\centering
\begin{minipage}[t]{0.395\textwidth}
\centering
\includegraphics[width=\linewidth]{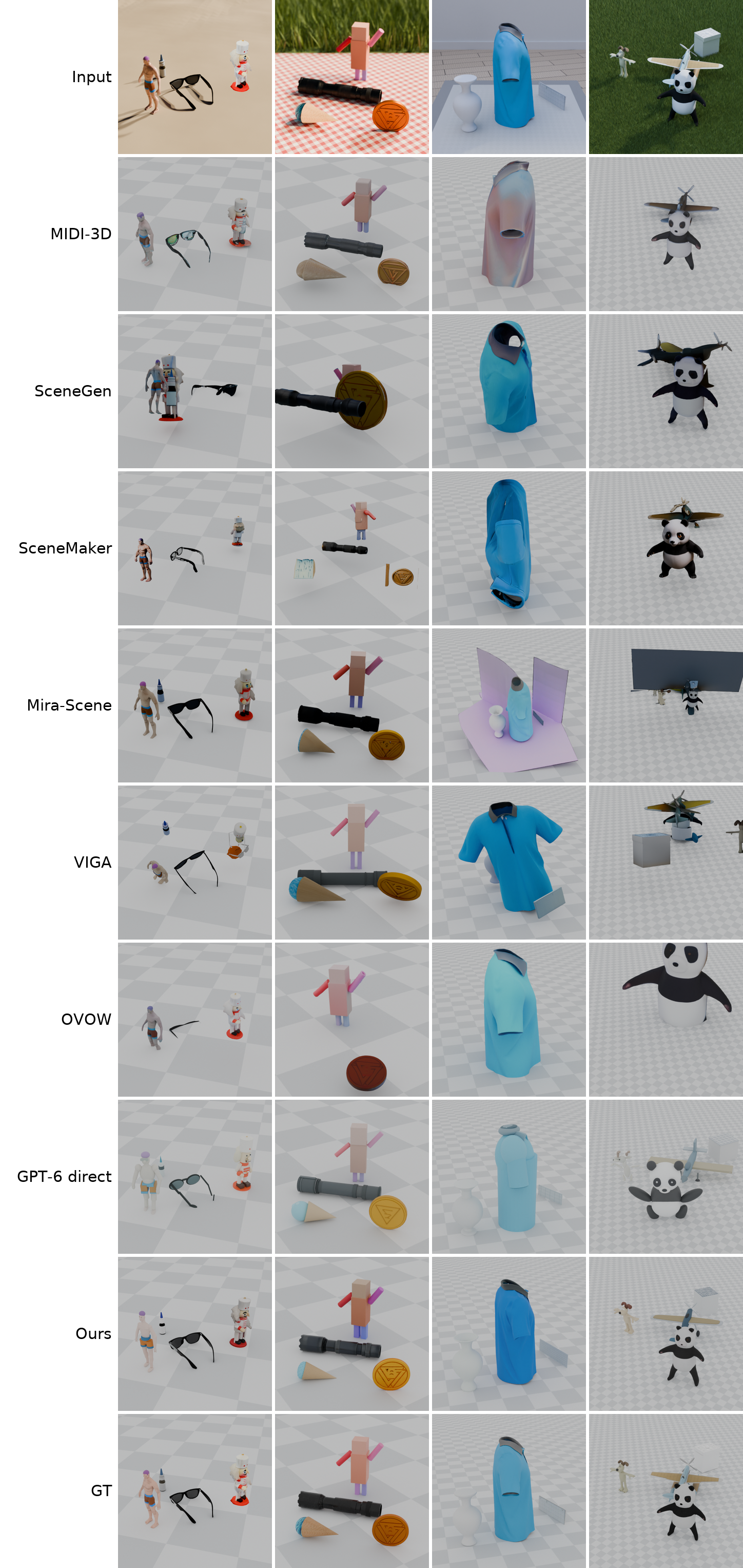}
\caption{Qualitative comparison on \benchthree{}, one scene per column. Predictions and ground truth are rendered from the ground-truth camera after the evaluator's alignment, on a common checkerboard ground without backgrounds, as \benchthree{} does not evaluate background reconstruction. All methods were evaluated on the benchmark input with the same ground-truth camera parameters.}
\label{fig:scene3d}
\end{minipage}\hfill
\begin{minipage}[t]{0.585\textwidth}
\centering
\includegraphics[width=\linewidth]{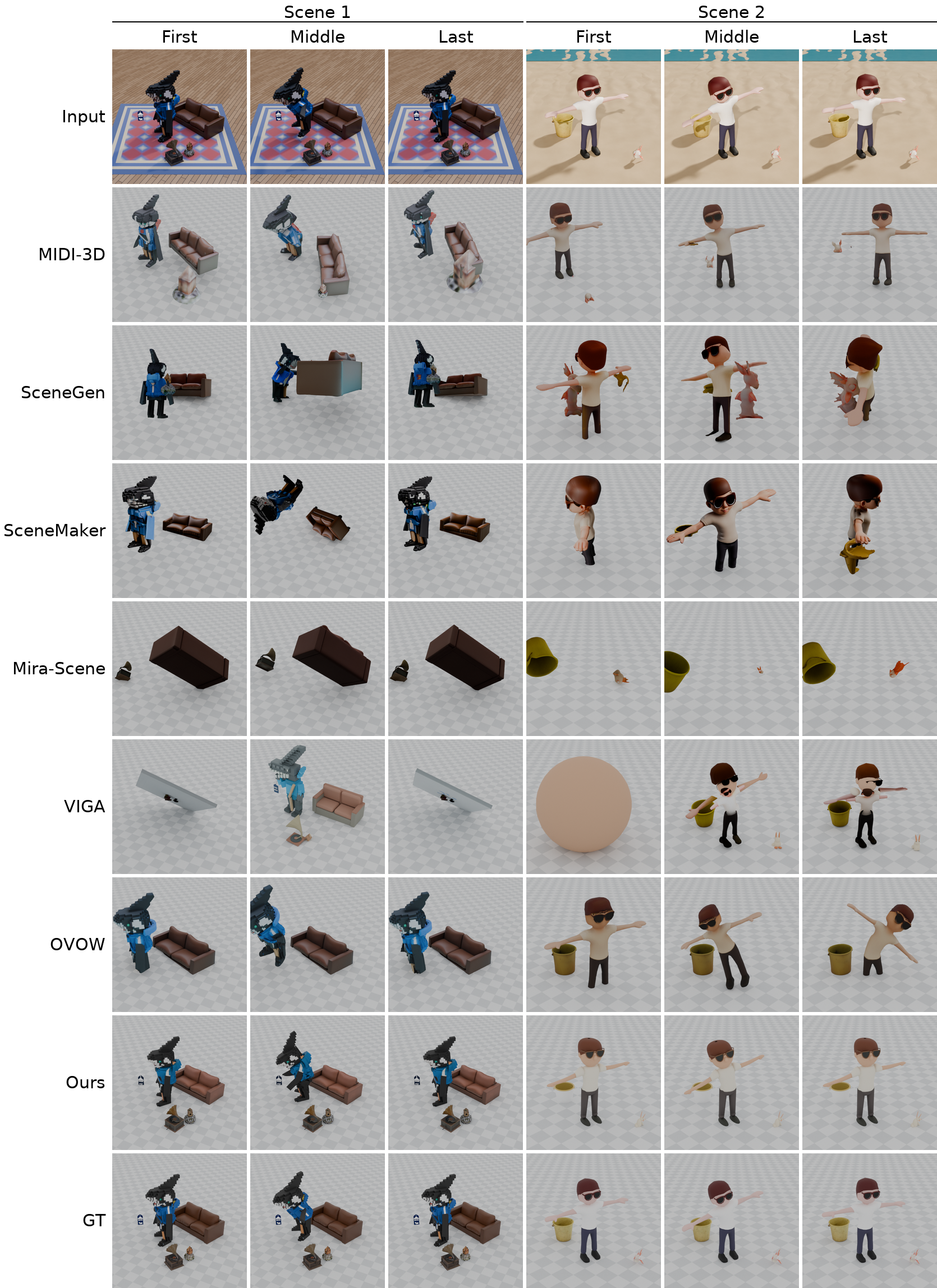}
\caption{Qualitative comparison on \benchfour{} at the first, middle, and last frames of two videos. MIDI-3D, SceneGen, SceneMaker, Mira-Scene, and VIGA are run on each frame independently. Predictions and ground truth are rendered from the ground-truth camera after the evaluator's alignment, on a common checkerboard ground without backgrounds, as \benchfour{} does not evaluate background reconstruction. For display, the inputs are re-rendered with a background that suits each scene; all methods were evaluated on the benchmark input with the same ground-truth camera parameters.}
\label{fig:scene4d}
\end{minipage}
\end{figure*}

\begin{figure*}[p]
\centering
\includegraphics[width=\textwidth]{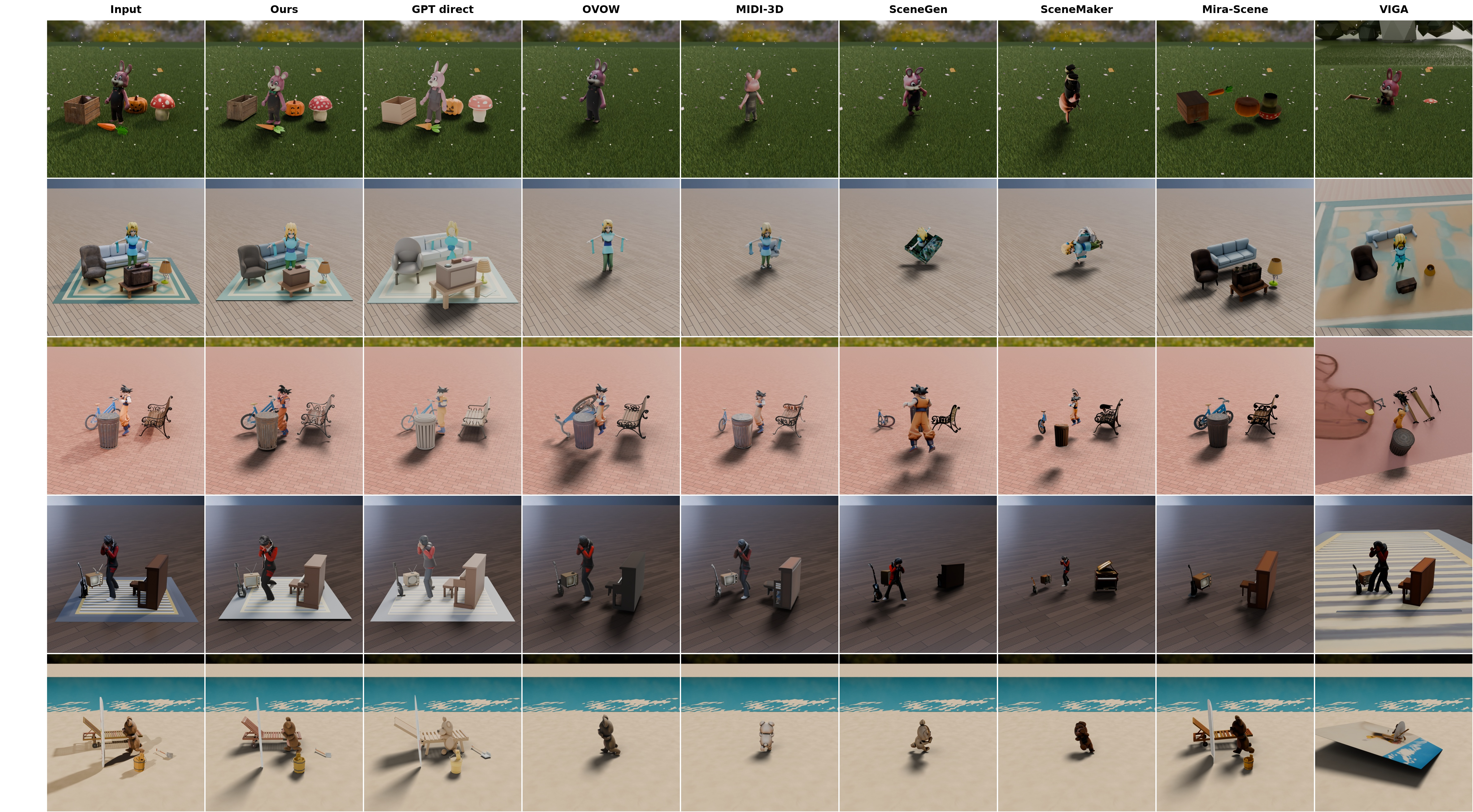}
\caption{3D comparison on the first frame of five demo videos outside the benchmark. Because existing baselines do not generate backgrounds, every method is rendered from the input camera with the same lights and shadows and composited onto the same background generated by our pipeline; baselines are placed by registering their outputs to the input camera without evaluator alignment.}
\label{fig:demo3d}
\vspace{14pt}
\includegraphics[width=0.70\textwidth]{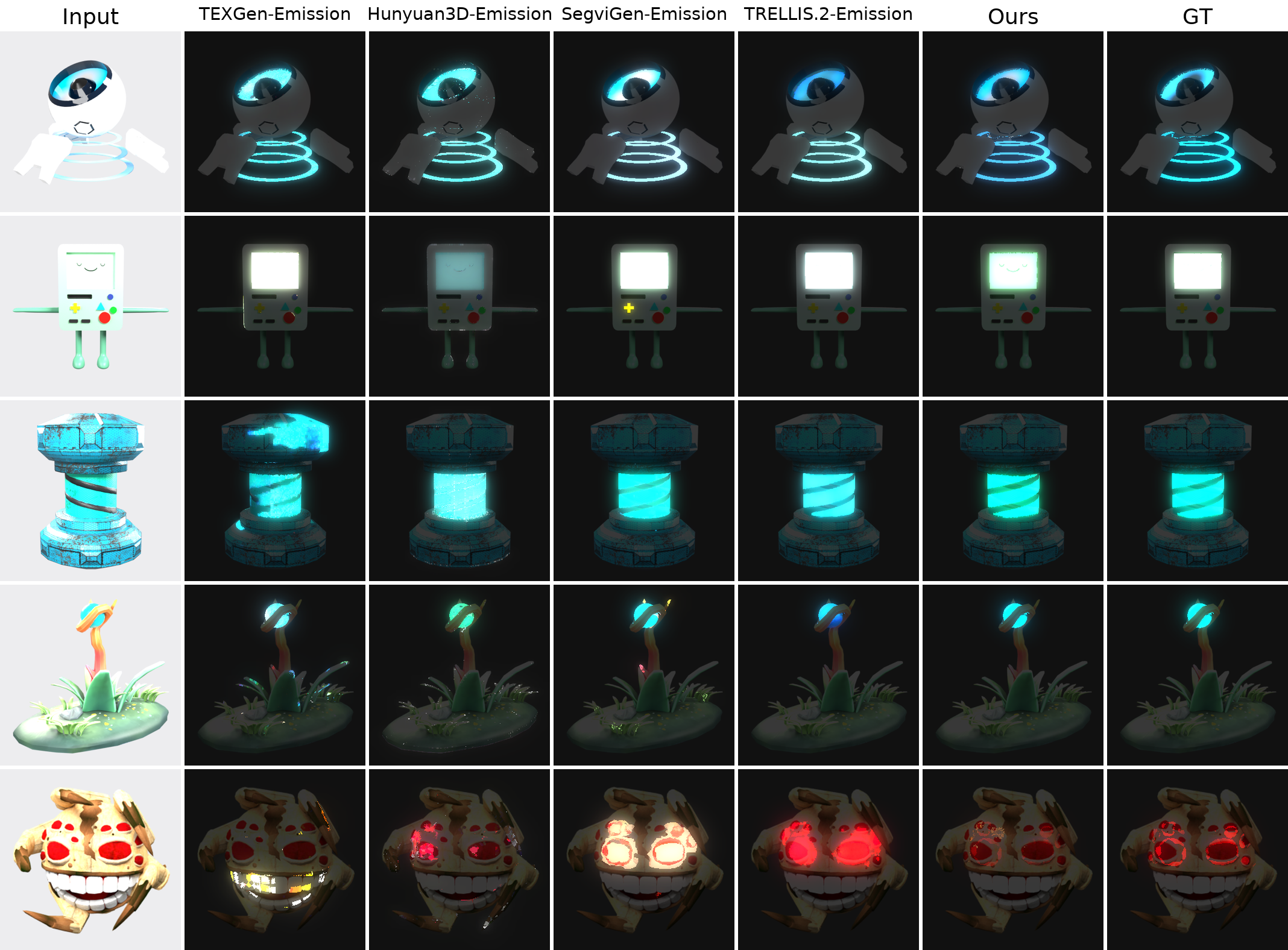}
\caption{Qualitative comparison of emission generation on LightGenBench test shapes. Each row shows one upright view: the input under studio lighting alongside each method's predicted emission overlaid on the darkened shape.}
\label{fig:emission}
\end{figure*}

\begin{figure*}[p]
\centering
\includegraphics[width=\textwidth,height=0.9\textheight,keepaspectratio]{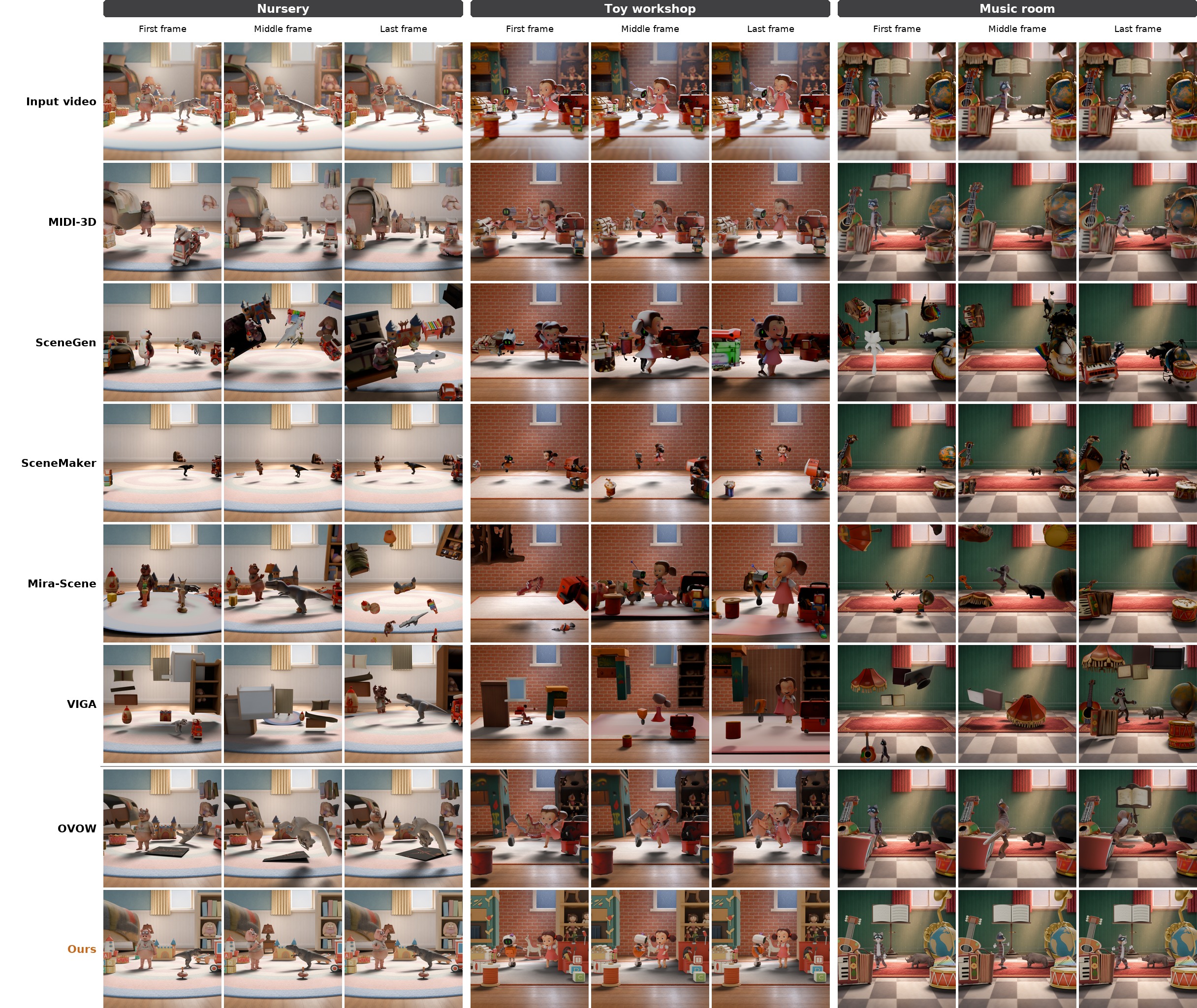}
\caption{4D comparison on the three animated-film videos of \cref{fig:demos} at their first, middle, and last frames. MIDI-3D, SceneGen, SceneMaker, Mira-Scene, and VIGA are run on each frame independently. Because existing baselines generally do not generate backgrounds, all methods are rendered from the input camera and composited onto the same background generated by our pipeline.}
\label{fig:film_baselines}
\end{figure*}

\Cref{fig:scene3d,fig:scene4d} show qualitative comparisons on \benchthree{} and \benchfour{}, respectively. By separating layout reasoning from object generation and preserving per-instance identity, \ours{} recovers object positions, orientations, and scales more faithfully than both compositional generators and direct agent baselines, while maintaining consistent static layouts and articulated character motion across video frames. On both \benchthree{} and \benchfour{}, all methods (both \ours{} and the compared baselines) which accept ground-truth camera parameters would use them; furthermore, \cref{fig:scene3d,fig:scene4d} omit backgrounds and render all outputs on a common checkerboard ground because these benchmarks evaluate foreground objects only and do not evaluate background reconstruction. \Cref{fig:demo3d,fig:film_baselines} extend these comparisons to out-of-benchmark scenes: the first frames of five synthetic demo videos and the three animated-film videos of \cref{fig:demos}, where per-frame single-image baselines suffer severe temporal flickering and missing objects whereas \ours{} produces coherent, editable 4D worlds. Because existing baselines generally do not generate backgrounds, in \cref{fig:demo3d,fig:film_baselines} we generate the background with our pipeline and composite every method's predicted objects onto that same shared background for visual comparison.

\section{Qualitative Emission Generation}
\Cref{fig:emission} shows qualitative examples from the LightGenBench test split. Compared with regression-based emission baselines that often miss fine light fixtures or spill emission onto dark regions, \ours{} localizes emissive boundaries cleanly across views and preserves the original surface colors by reusing base-color texels.

\section{Future Work}
We aim to extend the agentic framework studied here to rig and animate characters with arbitrary mesh topologies~\cite{long2026act,chen2026tailorrigcontrollablestructuredauxiliary}. We also plan to broaden scene generation and editing across indoor, outdoor, and traffic environments while preserving object identity and spatial relationships across edits~\cite{guo2026blenderforgeframeworkoptimizingreactive,xu2025cruisecooperativereconstructionediting,li2026genncap,yan20263dsceneeditorcontrollable3dscene,wang2026embodiedgenv2agenticsimulationready,he2026neoworldproprogramminginteractivescenes,xia2026sagescalableagentic3d,pfaff2026scenesmithagenticgenerationsimulationready,weng2026rfwwscalingphysicalasset,add_infinigen,add_holodeck,deitke2022procthorlargescaleembodiedai}. For simulation-ready assets, the agent could generate garment patterns, collision geometry, joints, and material parameters, then test their behavior in simulation~\cite{weng2026rfwwscalingphysicalasset,chen2026unigarment,lyu2026rfwaagenticproceduralgeneration,guo2026rfwseverydaydeformableobjects,wang2026manitwinscalingdatagenerationreadydigital,cao2026physxomniunifiedsimulationreadyphysical,zhou2026articraftagenticscalablearticulated,chen2024urdformerpipelineconstructingarticulated,xiang2020sapiensimulatedpartbasedinteractive}. For code-to-video generation, executable scene programs could specify geometry, motion, and camera paths for controllable rendering~\cite{chen2026videomodelsnative4d,chen2026codeworldmodelcoding,wang2026codeworldmodel20,chen2025code2videocodecentricparadigmeducational,chen2026code2videocompasscomprehensivebenchmarkgeneralpurpose,sensenovacodevideo,liao2025motionagentfinegrainedcontrollablevideo}. For games and film, such frameworks could generate scenes and production tasks, along with training data for interactive world models~\cite{liu2025lightxgenerative4dvideo,chen2026enginenativeeditable3dworld,xia2024video2gamerealtimeinteractiverealistic,bruce2024geniegenerativeinteractiveenvironments,valevski2025diffusionmodelsrealtimegame,zhang2025matrixgameinteractiveworldfoundation,he2026matrixgame20opensourcerealtime}. For autonomous driving, they could provide data and benchmarks for driving-scene reconstruction and generation, as well as closed-loop environments for training and evaluating driving policies~\cite{li2026genncap,wu2023marsinstanceawaremodularrealistic,hu2023planningorientedautonomousdriving,li2025unisceneunifiedoccupancycentricdriving,zhang2024droneassistedroadgaussiansplatting,ye2025gsocc3dscalingvisiononlyoccupancy,zhang2026plugandplaytrafficelementawareness,yang2024drivearenaclosedloopgenerativesimulation,gao2024vistageneralizabledrivingworld,zhao2024drivedreamer2llmenhancedworldmodels,yan2024drivingspherebuildinghighfidelity4d,Xiao_2026_CVPR}. For embodied AI, they could generate data for training reconstruction models and action-conditioned world models, and convert real scenes into simulation environments for reinforcement learning, policy evaluation, and real-to-sim-to-real transfer~\cite{xu2025diffusionknowstransparencyrepurposing,zhang2026dexoraopensourcevlahighdof,cui2026liberosafetycomprehensivebenchmarkphysical,chen2025robotwin20scalabledata,chen2026robodojounifiedsimandrealbenchmark,jang2025dreamgenunlockinggeneralizationrobot,kim2026cosmospolicyfinetuningvideo,nasiriany2024robocasalargescalesimulationeveryday,tao2025maniskill3gpuparallelizedrobotics,liu2023liberobenchmarkingknowledgetransfer,ranawaka2026simfoundrymodularautomatedscene,shuai2026gennav}.